\pdfoutput=1
\documentclass{CVM}
\usepackage{indentfirst}
\usepackage{verbatim}
\usepackage{longtable}
\usepackage{stfloats}
\usepackage{soul}
\usepackage{color, xcolor}
\usepackage[numbers,sort&compress]{natbib}
\soulregister{\cite}{7}
\soulregister{\ref}{7} 
\newcommand{\tabincell}[2]{\begin{tabular}{@{}#1@{}}#2\end{tabular}}

\begin{document}

\CVMsetup{
% -- type
type      = {Research/Review Article},
% -- doi
doi       = {s41095-0xx-xxxx-x},
% -- title
title     = {A survey on facial image deblurring},
% -- author
author    = {Bingnan Wang$^{1,2}$, Fanjiang Xu$^{1}$, and Quan Zheng$^{1}$ \cor{}\\
},
% -- runauthor
runauthor = {B. Wang, F. Xu, Q. Zheng},
% -- abstract
abstract  = {
  When a facial image is blurred, it significantly affects high-level vision tasks such as face recognition. The purpose of facial image deblurring is to recover a clear image from a blurry input image, which can improve the recognition accuracy, etc. However, general deblurring methods do not perform well on facial images. Therefore, some face deblurring methods have been proposed to improve performance by adding semantic or structural information as specific priors according to the characteristics of the facial images. In this paper, we survey and summarize recently published methods for facial image deblurring, most of which are based on deep learning. First, we provide a brief introduction to the modeling of image blurring. Next, we summarize face deblurring methods into two categories: model-based methods and deep learning-based methods. Furthermore, we summarize the datasets, loss functions, and performance evaluation metrics commonly used in the neural network training process. We show the performance of classical methods on these datasets and metrics and provide a brief discussion on the differences between model-based and learning-based methods. Finally, we discuss the current challenges and possible future research directions.
},
% -- keywords
keywords  = { Facial image deblurring,  Model-based, Deep learning-based, Semantic or structural prior},
% -- copyright
copyright = {The Author(s)},
}

\maketitle

    \begin{figure}[b] \vskip -4mm
    \small\renewcommand\arraystretch{1.3}
        \begin{tabular}{p{80.5mm}} \toprule\\ \end{tabular}
        \vskip -4.5mm \noindent \setlength{\tabcolsep}{1pt}
        \begin{tabular}{p{3.5mm}p{80mm}}
    $1\quad $ & Institute of Software, Chinese Academy of Science, Beijing 100190, China. E-mail:  fanjiang@iscas.ac.cn; zhengquan@iscas.ac.cn\cor{}.\\
    $2\quad $ & University of Chinese Academy of Sciences, Beijing 100049, China. E-mail: wangbingnan21@mails.ucas.ac.cn.
   \\
&\hspace{-5mm} Manuscript received: 2022-08-17; accepted: 2023-01-19 \vspace{-2mm}
    \end{tabular} \vspace {-3mm}
    \end{figure}

\section{Introduction}

\begin{figure}[htbp]
	\centering
	\subfigure[Sharp image]{
		\centering
		\includegraphics[width=0.3\linewidth]{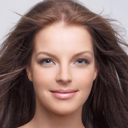}
	}%
	\subfigure[Motion blur]{
		\centering
		\includegraphics[width=0.3\linewidth]{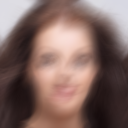}
	}%
	\subfigure[Defocus blur]{
		\centering
		\includegraphics[width=0.3\linewidth]{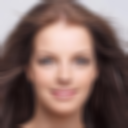}
	}%
	
	\centering
	\caption{Different types of blurry facial images.}
	\label{fig1}
\end{figure}

Facial image deblurring is a technique used to recover clear facial images with sharp textural details from blurry facial images. Blurry images are widespread in life, which can be caused by various reasons, such as optical aberrations, camera shake, and object movement. The two most common types of blur are motion and defocus blur.
An example of blurry facial images caused by these two types is shown in Fig.~\ref{fig1}. 

With the development of technologies such as face recognition, the processing of degraded facial images has become an important research topic. In surveillance or on some open occasions, the faces in the images are prone to motion blur, which can greatly reduce the performance of systems such as face recognition and video surveillance. Therefore, facial image deblurring has become a significant research topic in computer vision, and an increasing number of researchers have conducted studies on facial image deblurring.

\subsection{Problem definition}

Blur can be caused by various reasons, and we can represent it by the unified blur operator $K$. Moreover, considering that there may be noise in the image degradation process, we use the following model to represent the degradation process:
\begin{equation}
	\label{eq1}
	Y=D(X, K, n)
\end{equation}

Here, $Y$ refers to the degraded image, $X$ refers to the clear image, $K$ refers to the blur operator, and $n$ refers to the noise. In the degradation process, we consider noise to be additive. In addition, many methods simplify the blur degradation process to a linear function and assume that the blur in an image is spatially invariant; therefore, Eq. ~\eqref{eq1} can be rewritten as the following simplified model:
\begin{equation}
	\label{eq2}
	Y=k * X+n
\end{equation}

where $k$ refers to the blur kernel and $*$ refers to the convolution operation. Image deblurring is used to obtain a clear image $X$ from a blurry image $Y$. According to the different blurring causes, the blur kernel $k$ can be modeled in different forms, as described below.

\textbf{Defocus blur} \cite{lu2017out}. When the object is not in the focal plane of the camera or the scene has a short depth of field, unfocused areas can produce unclear details and textures. The point spread function (PSF) of out-of-focus blur can be represented by the following model:
\begin{equation}
	\label{eq3}
	k(i, j)= \begin{cases}\frac{1}{\pi r^{2}} & \text { if } i^{2}+j^{2} \leq r^{2} \\ 0 & \text { otherwise }\end{cases}
\end{equation}

where $r$ is the radius of the blur and $(i,j)$ refers to the pixel coordinates. 

\textbf{Motion blur.} This is caused by the motion of the camera or the movement of the object. When the camera moves in a fixed direction at the moment of shooting, the blur kernel can be modeled using Eq. ~\eqref{eq4}, where $L$ is the motion length and $\theta$ is the motion angle. However, actual motion blur is more complicated than this one. On the one hand, the direction and degree of movement of the camera can vary; on the other hand, in a dynamic scene, only some objects move and others are stationary. Several methods have been proposed to generate simulated blur  kernels \cite{chakrabarti2016neural,boracchiModelingPerformanceImage2012}. However, how to ensure the diversity of the simulation kernels and the authenticity of the distribution is still a problem.

\begin{equation}
	\label{eq4}
	k(i, j)= \begin{cases}\frac{1}{L} & \text { if } \sqrt{i^{2}+j^{2}} \leq \frac{L}{2} \text { and } \frac{i}{j}=-\tan \theta \\ 0 & \text { otherwise }\end{cases}
\end{equation}

\textbf{Gaussian blur.} Many methods use a simple Gaussian function in the model to represent the blurring process \cite{song2019joint,wang2021towards}. The Gaussian kernel function can be represented by the following model: 
\begin{equation}
	\label{eq5}
	k(i, j)=\frac{1}{2 \pi \sigma^{2}} e^{-\frac{i^{2}+j^{2}}{2 \sigma^{2}}}
\end{equation}

where $\sigma$ is the standard deviation that indicates the degree of blur.

\subsection{Scope of this survey}

Facial image deblurring is a domain-specific image deblurring problem. The corresponding solution is improved and developed by using general deblurring methods. Some reviews \cite{li2022survey,zhang2022deep} summarize the existing general deblurring models. Li \cite{li2022survey} provided a brief summary of traditional and depth-represented image deblurring methods, whereas Zhang et al. \cite{zhang2022deep} focused on the detailed introduction of   deep learning-based image deblurring methods. 

In contrast to these studies, this survey focuses on summarizing the deblurring research conducted on facial images. As a special application scenario, there are fewer textures and edges on facial images than those on general scene images; therefore, the proposed general deblurring methods cannot produce good results on facial images. In addition, although the identity is changed, different faces are composed of fixed components that can be used as prior information to improve the performance of the general methods. Based on these, a number of studies have been conducted dedicated to the deblurring of facial images, which we summarize in this article.

\begin{figure*}[h!t]
	\centering
	\includegraphics[width=.7\linewidth]{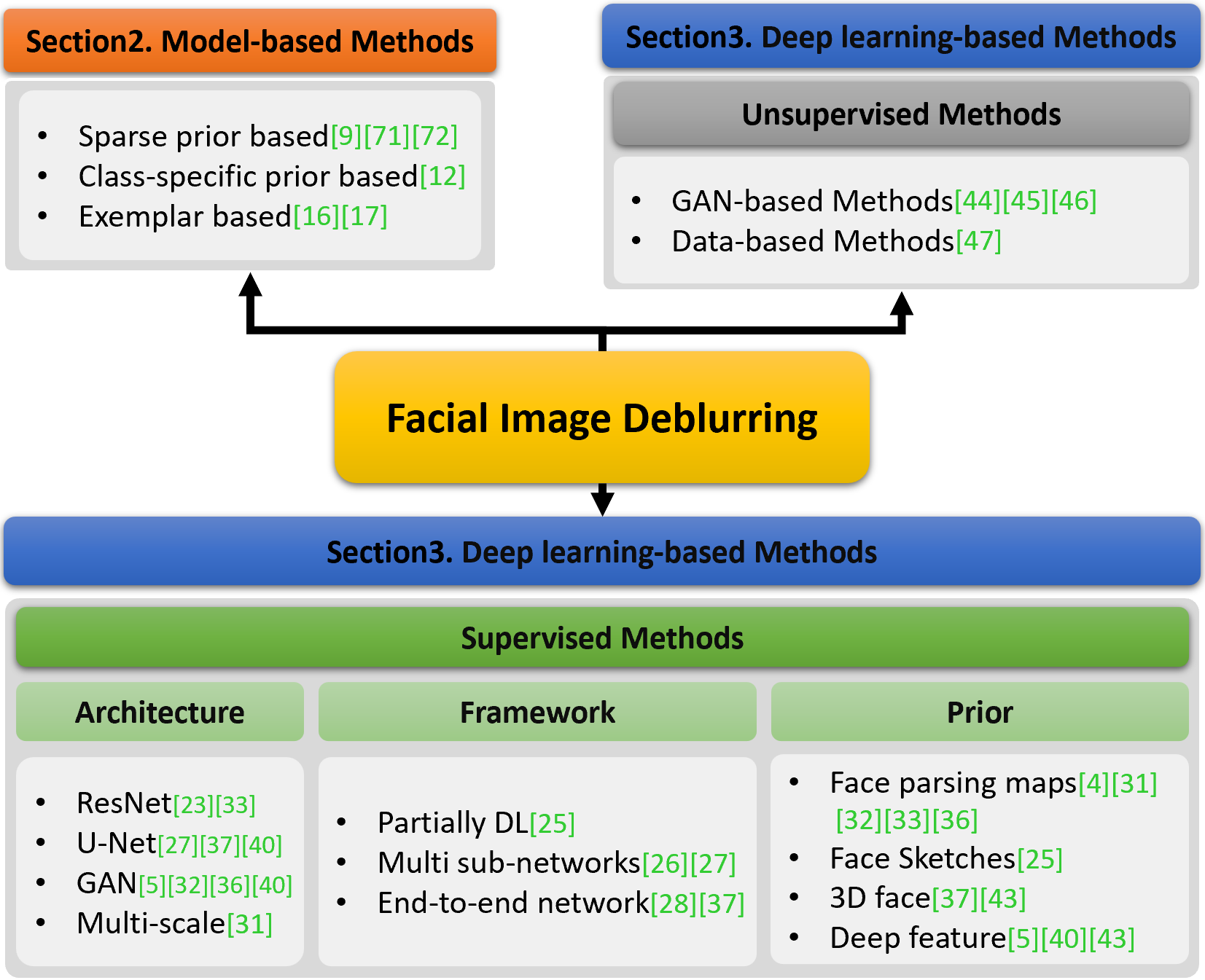}
	\caption{Taxonomy of the facial image deblurring methods studied in our paper. ``DL'' indicates ``deep learning''. Details of these methods can be found in Sections 2 and 3.}
	\label{fig2}
\end{figure*}

Before 2015, the methods used for face deblurring were mainly model based. Some methods have been proposed to improve the recognition performance, whereas others improve general image deblurring methods in the spatial domain. Here, we mainly summarize the research conducted in the field of face deblurring after 2010. Since 2016, owing to the strong fitting ability of a convolutional neural network, methods based on deep learning have been gradually proposed and have achieved better performance. Therefore, we mainly introduce the deep learning-based methods in this survey. In the deep learning model, different methods aim to introduce sufficient priors into the model to alleviate the problem of fewer facial image textures. The taxonomy of the methods involved is shown in Fig.~\ref{fig2}.

The rest of the paper is structured as follows. We will provide a brief introduction to the model-based methods in Section 2. Then, we summarize the learning-based methods in Section 3. Section 4 introduces the commonly used blurry-sharp facial datasets in learning-based methods. Section 5 lists the loss functions and neural network training strategies commonly used in various models. Section 6 lists the metrics used to evaluate the image quality. In Section 7, we compare the performance of existing typical methods (including the model-based and deep learning-based methods).   Section 8 provides a brief discussion and a macro comparison of the main differences between the model-based and learning-based methods. Finally, in Section 9, we summarize the limitations of the current study and the future research directions.

\section{Model-based methods}
The earliest methods transform a blurry image into the frequency domain to solve this problem. Nishiyama et al. \cite{nishiyama2010facial} addressed the blur degradation problem by learning the feature subspace of facial images in the frequency domain. Blurry images with the same PSF are projected into the same subspace, indicating that they have the same degree of blurring. At the inference time, the kernel of the blurry image is determined as the PSF of the nearest subspace. However, these methods can only deal with fixed blur degradation and cannot be generalized to real-world blurry images with complicated and nonuniform degradation.

Most model-based methods focus on solving problems in the spatial domain. According to the degradation model established by Eq. ~\eqref{eq2}, obtaining clear facial images can be expressed as

\begin{equation}
	\label{eq6}
	x, k=\underset{x, k}{\arg \min }\left[\|k * x-y\|_{2}^{2}+f(x)+g(k)\right]
\end{equation}

where the first term on the right-hand side of the equation is the fidelity term; $x$ and $y$ denote latent sharp images and blurry images, respectively; $*$ denotes the convolution operation; and $f(x)$ and $g(k)$ are the regularization terms for the latent sharp image and the blur kernel, respectively. Because it is an ill-posed problem to simultaneously estimate the blur kernel and the sharp image from a blurry image, adding a regularization term is necessary to narrow the solution space. Eq.~\eqref{eq6} is usually solved iteratively, that is, by estimating the blur kernel and recovering the latent image iteratively. The main idea of traditional image deblurring methods is to restore salient edges implicitly or explicitly to estimate blur kernel $k$. However, in facial images, only a few edges are available for blur kernel estimation. Therefore, existing general deblurring methods cannot achieve satisfactory results for blurry facial images.

Zhang et al. \cite{zhang2011close} adopted the sparse prior for the regularization of the latent image and the $L_2$ norm prior for the regularization of the blur kernel. Moreover, a sparse representation prior was added to the face recognition process, and then clear facial images were reconstructed by jointly optimizing the process of face restoration and recognition. By combining these two tasks, the authors were able to demonstrate a significant improvement over treating them individually. However, this method is only effective for facial images with good face alignment and simple motion blurring. Given a specific patch of facial images, there are many similar non-local patches near it. Using this feature, Tian et al. \cite{tian2016weighted} introduced the earliest weighted non-local self-similarity \cite{jiang2014mixed} method for denoising into a sparse representation model and verified its effectiveness in face deblurring. Subsequently, Anwar et al. \cite{anwar2018image} proposed class-specific priors by transforming the images of a specific class into a Fourier space. Specifically, they learned a subspace spanned by the filter responses of sharp images in each class to a bandpass filter. In this manner, the method achieves improved results when dealing with blurry images lacking high-frequency details. 

Zhang et al. \cite{zhang2022pixel} found that, in the iterative optimization process, some pixels of the intermediate restored image did not satisfy the model of Eq. ~\eqref{eq2}, which was unfavorable for the next kernel estimation process. They proposed a pixel screening method to correct intermediate images and screen out bad pixels to facilitate more accurate kernel estimation. Tian and Tao \cite{tian2015coupled} updated Eq. ~\eqref{eq2} by redefining the blur kernel and latent sharp images. They represented the PSF as a linear combination of a set of predefined orthogonal PSFs. Similarly, the estimated intrinsic (EI) sharp facial image was represented as a linear combination of a set of predefined orthogonal facial images. The coefficients of PSF and EI were jointly learned by minimizing the reconstruction error. Finally, they used a blind image quality assessment\cite{mittal2012making} method to automatically select the best images.

\begin{figure}[htbp]
	\centering
	\subfigure[Input]{
		\centering
		\includegraphics[width=0.3\linewidth]{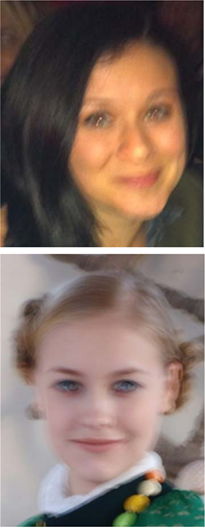}
	}%
	\subfigure[Exemplar image]{
		\centering
		\includegraphics[width=0.3\linewidth]{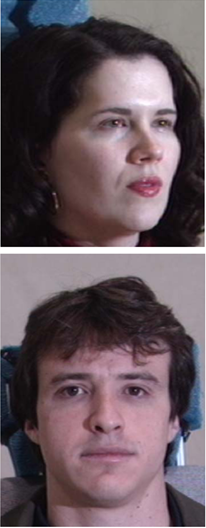}
	}%
	\subfigure[Predicted $\nabla S$]{
		\centering
		\includegraphics[width=0.3\linewidth]{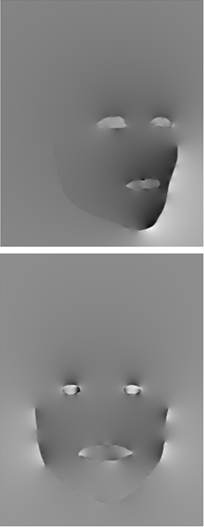}
	}%
	
	\centering
	\caption{Facial structures extracted from blurry input images by Pan et al. \cite{pan2014deblurring}. (a) Blurry facial image to be restored; (b) matched example facial image; (c) corresponding edge of the facial structure.}
	\label{fig3}
\end{figure}

In addition, some methods using reference images have been developed. Hacohen et al. \cite{hacohenDeblurringExampleUsing2013} proposed a facial image deblurring method that uses a reference image, which has the same scene as that of the original blurry image. The function of the reference image is twofold. On the one hand, information from reference images can facilitate the process of kernel estimation. On the other hand, it can serve as a strong local prior for nonblind deconvolution. The algorithm performs well in deblurring certain classes of images. However, the use of the same reference image as the input observation has certain limitations in practical applications. Subsequently, Pan et al. \cite{pan2014deblurring} proposed a new method that does not require the sample and test images to have the same identity and background. The authors constructed an example dataset of facial images from the CMU PIE dataset and extracted important structures for each image, including facial contours, eyes, and mouths. During the test, the test image was compared with the structure of the sample and the best match was found. Fig.~\ref{fig3} shows an example of the matched facial structures given in the paper. This structure is used to recover the image edges and guide the blur kernel estimation process. After the face edge is obtained, also according to Eq.~\eqref{eq2}, the blur kernel is estimated alternately and the latent image is restored. The face structure information in this exemplar library can help extract face edges and eliminate the phenomenon of ringing and artifacts in traditional edge selection methods. However, matching each image to an example dataset is computationally intensive. Processing a single image takes several hours. Moreover, this type of approach tends to be inaccurate for face poses and angles that are not present in the dataset.

\section{Methods based on deep learning}
According to Zhang et al. \cite{zhang2022deep}, we can conclude that there are some commonly used basic blocks in neural networks for image deblurring, such as a convolution block, ResBlock, Inception block \cite{szegedy2015going}, and DenseBlock. In addition, there are some commonly used network architectures such as U-Net \cite{ronneberger2015u}, multi-scale networks \cite{nah2017deep}, generative adversarial networks (GANs) \cite{goodfellow2014generative}, and cascade networks \cite{schuler2015learning}. U-Net consists of an encoder, a decoder, and skip connections that can perform image transformation in an end-to-end manner. The multi-scale network feeds the original size image and the downsampled low-resolution image into the network. It first performs image restoration on a small scale and then restores the images up to their original sizes. It performs image deblurring in a coarse-to-fine manner, which greatly increases the computational load of the model. Similarly, cascaded networks can generate higher quality images owing to the concatenation of the networks. GANs can generate diverse and realistic facial images; however, the disadvantage is that sometimes the generated sharp facial image is not identical to the corresponding blurry image.

To reduce the ill-posedness of the problem, Jin et al. \cite{jinLearningFaceDeblurring2018} improved the basic block. They advocated the expansion of the range of receptive fields in ResNet \cite{he2016deep}. They employed a resampling convolution operation that ensured a wide receptive field (RF) from the first layer while being computationally efficient. The advantages of using a large receptive field can be analyzed in two ways. First, the last few layers of the network can take advantage of higher abstract levels to better understand the image content and ultimately produce better deblurred outputs. Second, a larger RF contains more structural information. Moreover, more erroneous latent image-kernel pairs can be excluded compared with using a small RF with less structure, thereby reducing the ill-posedness of the system. To avoid artifacts, the authors concatenated a hybrid subnetwork and a deblurring network by applying several convolutional layers to process local images.

Many models use a combination of the above network architectures and basic blocks to achieve good restoration of blurry faces. Next, we will summarize the deep learning methods of facial image deblurring separately according to whether paired training data are required. Additionally, we provide a summary of these methods in Table \ref{tab4}.
\subsection{Supervised learning}
\begin{figure*}[h!t]
	\centering
    \includegraphics[width=.8\linewidth]{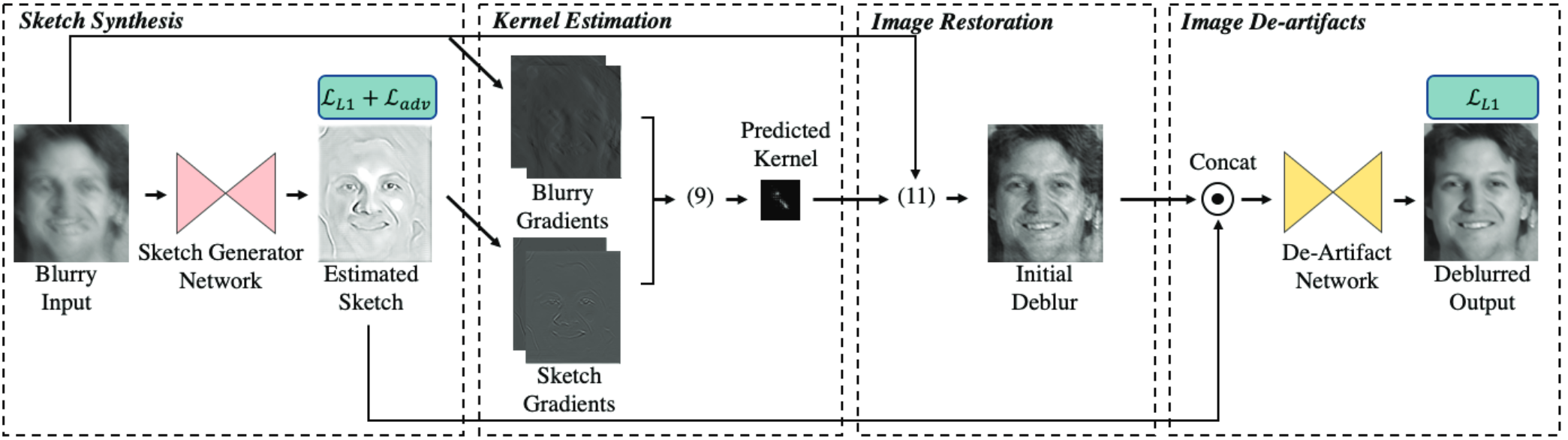}
	\caption{Example of the first class of methods \cite{linLearningDeblurFace2020}. The blurry input first goes through a neural network to generate an estimated sketch. The traditional optimization processes of kernel estimation and deconvolution are then performed on the basis of the gradients of the sketch and blurry input; finally a clear output is obtained through the de-artifacts network.}
	\label{fig4}
\end{figure*}

Some deep learning methods have been developed for image deblurring and applied to facial images. These methods can be roughly divided into three categories: The first class of methods first utilizes a neural network to estimate the blur parameters and then restores sharp images in model-based image deblurring frameworks. Chakrabarti \cite{chakrabarti2016neural} utilized a convolutional neural network (CNN) to compute the complex Fourier coefficients of the deconvolution filters of each image patch, followed by nonblind deconvolution. Lin et al. \cite{linLearningDeblurFace2020} utilized U-Net to extract face sketches for blur kernel estimation and nonblind deconvolution. Its architecture is shown in Fig.~\ref{fig4}.

\begin{figure*}[h!t]
	\centering
	\includegraphics[width=.8\linewidth]{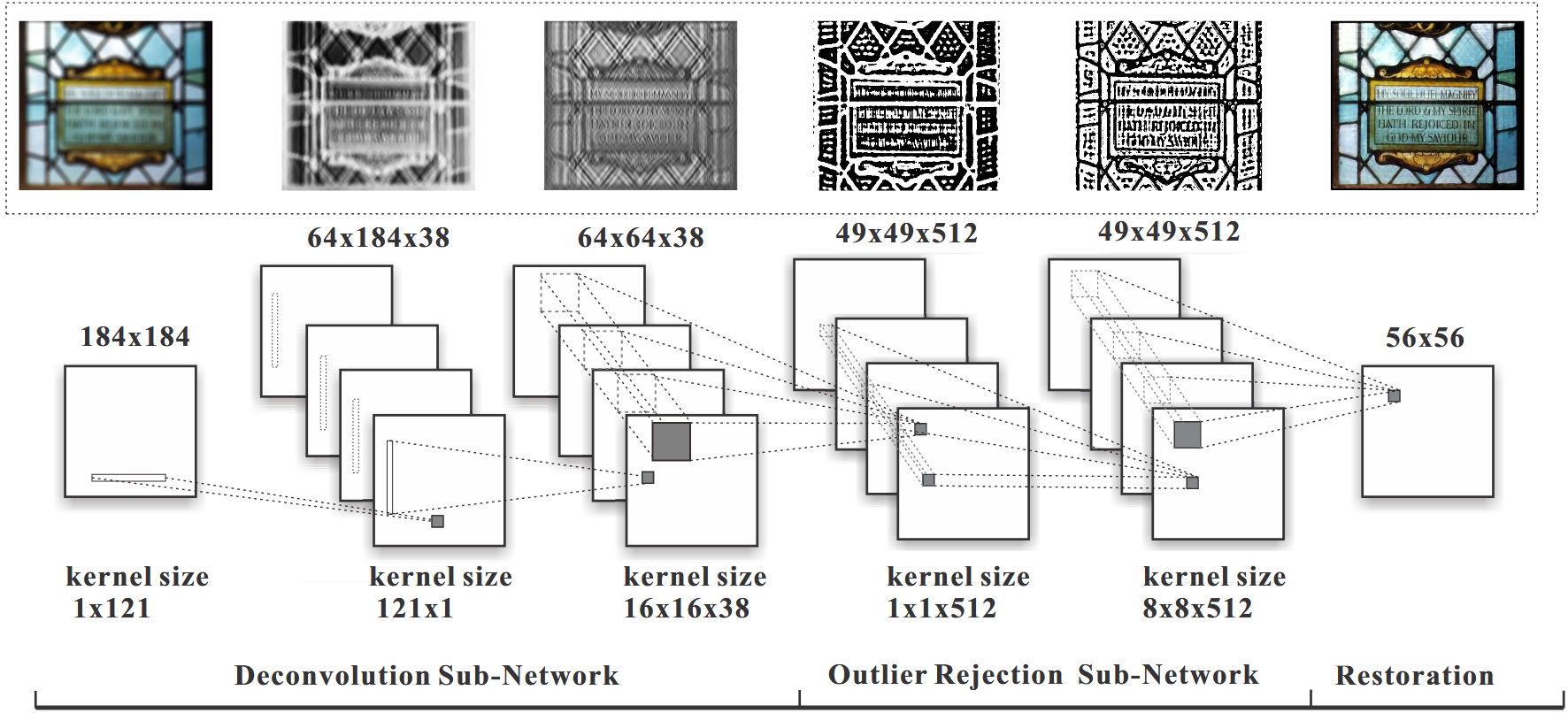} 
	\caption{Example of the second class of methods \cite{xu2014deep}. This model performs deconvolution and de-artifacting sequentially through two network modules.}
	\label{fig5}
\end{figure*}

The second class of methods uses multiple subnetworks to improve the deblurring results. Schuler et al. \cite{schuler2015learning} used a cascaded network structure to iteratively perform the kernel estimation and deconvolution processes. The algorithm adopted a coarse-to-fine strategy similar to that used in model-based deblurring methods. However, the network does not generalize well to different and diverse kernel sizes. Xu et al. \cite{xu2014deep} constructed two subnetworks for image deconvolution and artifact removal. Their architectures are shown in Fig.~\ref{fig5}.  Chrysos et al. \cite{chrysosMotionDeblurringFaces2019} adopted a two-step architecture to perform facial image deblurring. The first step uses a high-performance hourglass network to recover the low and medium frequencies of the image. The second step recovers the high-frequency details of the images by training a conditional GAN while ensuring that the output images are close to natural images.

\begin{figure}[h!t]
	\centering
	\includegraphics[width=1\linewidth]{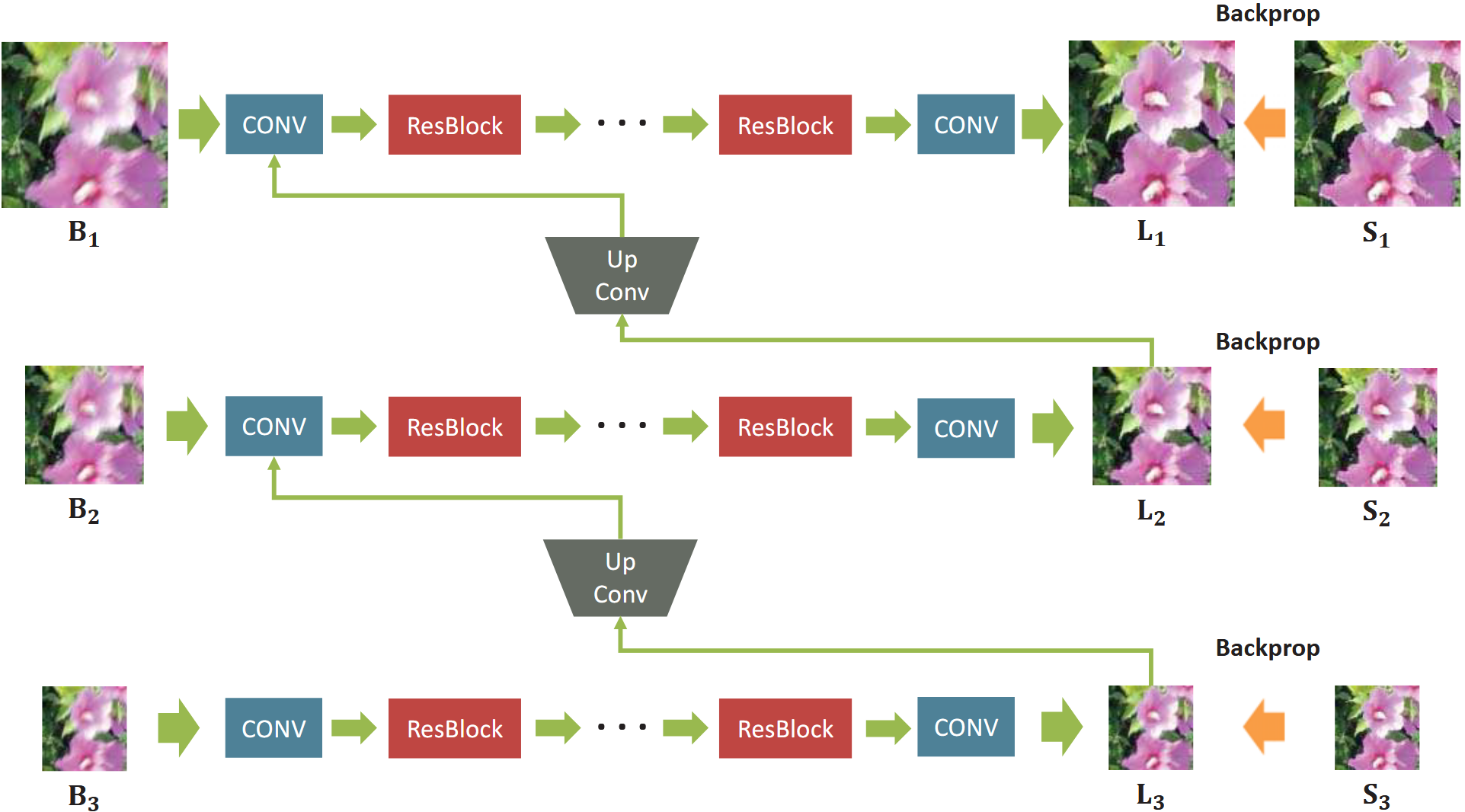}
	\caption{Example of the third class of methods \cite{nah2017deep}, which is the end-to-end deblurring neural network. This method first recovers the low-resolution blurry image $B_3$ and then upsamples the result into the deblurring framework of higher-scale $B_2$. Finally, the original scale image $B_1$ is restored.}
	\label{fig6}
\end{figure}

The third class of methods uses an end-to-end learning approach to model the entire restoration process. Nah et al. \cite{nah2017deep} proposed a multi-scale CNN to directly perform image deblurring without an intermediate kernel estimation step. The architecture of their model is shown in Fig.~\ref{fig6}. Chrysos et al. \cite{chrysosDeepFaceDeblurring2017} were the first to explicitly use deep architectures for face deblurring. They improved the classic ResNet architecture to perform end-to-end face deblurring tasks. In these architectures, face alignment techniques are used to preprocess each face and weak supervision is applied to exploit the structures of the faces. However, during testing, preprocessing of the blurry input is prone to errors, resulting in poor subsequent deblurring. Wang et al. \cite{wang2017deepdeblur} stacked multi-scale Inception  modules in a residual manner to perform face deblurring. The convolution kernels of different sizes of the Inception \cite{szegedy2015going} module can deal with different blurring degrees of the input image; however, they are memory intensive. Qi et al. \cite{qiBlindFaceImages2021} employed a GAN architecture to explore its specific effects on face deblurring. The authors adopted an improved U-Net and a feature enhancement module as the generator. Through careful design of the basic network blocks, enhanced feature representation, and adversarial training, the proposed method can generate more realistic faces.

There are a few edges and textures in the facial images. When we only input the blurry-sharp face pairs to the model, the output deblurred facial images are not sufficiently clear, with artifacts and poor details. Based on the above three types of methods, many methods have explored the effectiveness of prior information in the face, such as semantics, for deblurring tasks. Although the different faces are very multifarious, they are composed of fixed components. To fully use the face features, many methods focus on extracting specific information from facial images as a prior to guide the network training. The priors proposed by recent studies are facial geometry priors (such as face landmarks \cite{chrysosDeepFaceDeblurring2017}, face parsing maps \cite{song2019joint,shenDeepSemanticFace2018,yasarla2020deblurring,shenExploitingSemanticsFace2020,leeProgressiveSemanticFace2020}, and 3D facial models \cite{renFaceVideoDeblurring2019,zhu2022blind}) and deep feature priors \cite{jungDeepFeaturePrior2022,wang2021towards,zhu2022blind}. 

\begin{figure*}[h!t]
	\centering
	\includegraphics[width=.7\linewidth]{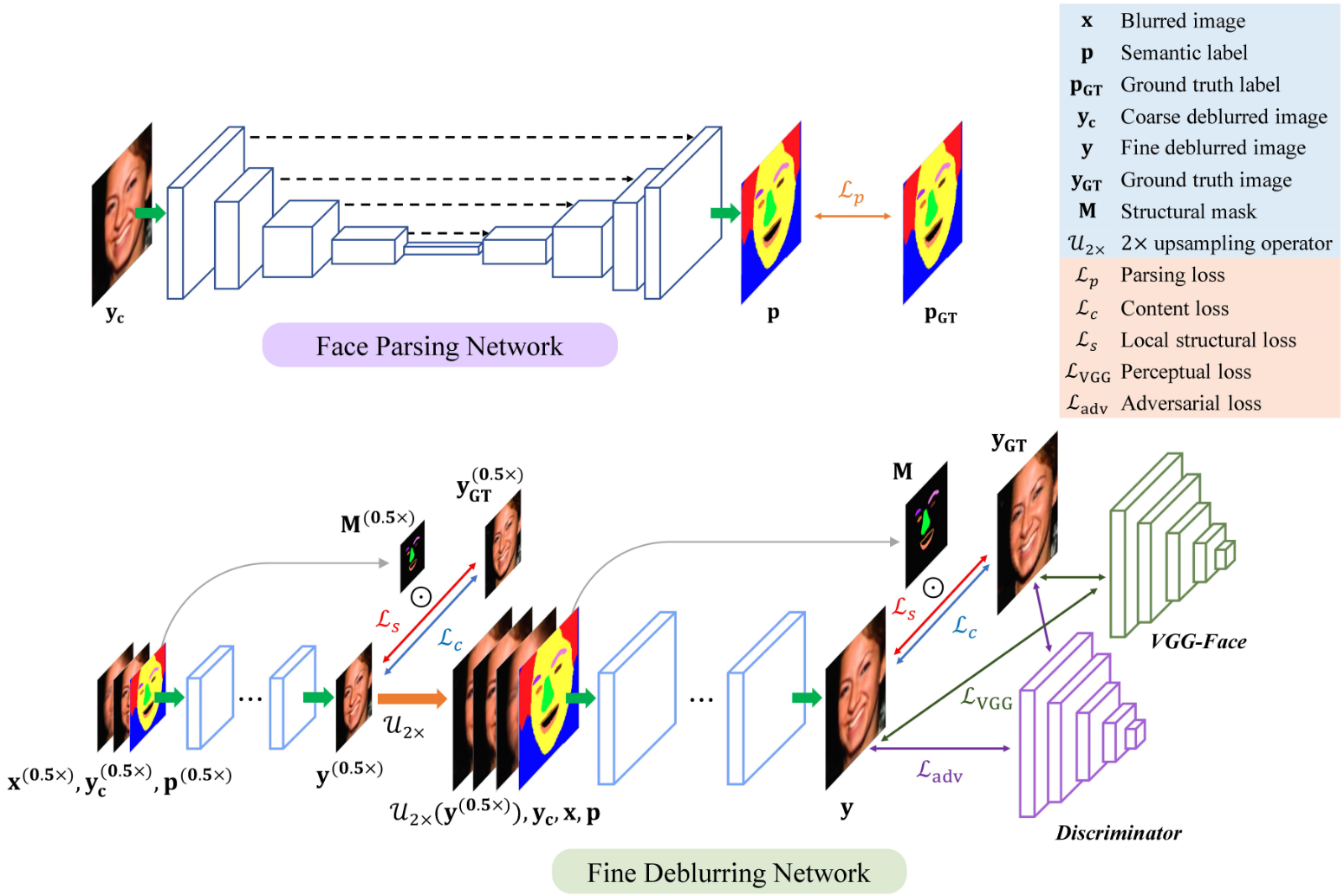}
	\caption{Use of a face semantic map as a prior for the deblurring neural network \cite{shenExploitingSemanticsFace2020}. This method first obtains the semantic labels of blurry facial images through a face parsing network. Then, the semantic map is concatenated with the blurry image in the channel dimension as the input of the deblurring network.}
	\label{fig7}
\end{figure*}

Shen et al. \cite{shenDeepSemanticFace2018} first attempted to incorporate the semantic parsing information of facial images into a network as a prior. The architecture of their model is shown in Fig.~\ref{fig7}, which is a typical model with a face semantic map as a prior. The authors adopted the deblurring network proposed by Nah et al. \cite{nah2017deep}. They obtained the semantic map of the blurry facial images through the semantic segmentation network and concatenated the semantic map and the blurry image in the channel dimension as the input of the deblurring network. To enhance the local performance, such as on the eyes, eyebrows, and mouth, Shen et al. also proposed a local structure loss to constrain the local output of the network. However, when there is severe motion blur in the input image, the semantic maps extracted from the blurry image are likely to be incorrect. Consequently, the authors improved their work in 2020 \cite{shenExploitingSemanticsFace2020}. To extract the correct face parsing map, Shen et al. added a coarse deblurring network before the face parsing network to reduce the blur in the input image. Then, the face parsing network extracts the semantic maps from the coarse deblurred images. Finally, the fine deblurring network recovers sharp facial images from the given blurry input images, coarse deblurred images, and corresponding semantic maps. Another improvement is that, in the local structure loss of the face components, different weights are set for each face component separately, instead of using fixed weights for all components. This adaptive local structure loss can adjust the weights and recover the fine details on the basis of the size of each facial component.

In the same year as the work of Shen et al. \cite{shenDeepSemanticFace2018}, Song et al. \cite{song2019joint} exploited semantic information to jointly perform face super-resolution and deblurring. The model mainly consists of two modules. The first module is a deep CNN called the facial structure generation network. The network takes the degraded image and semantic masks as inputs and predicts a base image containing the basic structure of the face. To enhance the facial details in the base image, the authors developed a detailed enhancement algorithm using high-resolution (HR) example images. In the first step, the patch correspondence between the base image and the HR example image is established. Then, the base image is regressed using the patches from the HR example image and an intermediate result is obtained. In the second step, the details of the intermediate result are passed to the base image through an edge-preserving filter to obtain the final result. The establishment of the detail enhancement module ensures that, even if the semantic mask extraction of the input blurry image is inaccurate, a real face with detailed edges can be recovered.

However, some facial components, such as the eyes, eyebrows, and mouth, cannot be reconstructed well because of the small face area. Yasarla et al. \cite{yasarla2020deblurring} noticed a class imbalance in the semantic map of faces. They proposed a confidence measure-based multi-stream semantic network. In the first stage of the network, different streams are used to complete the semantic map feature extraction of each class. The extracted features are concatenated together and fed into the second stage and used as input together with the semantic map fusion. In the second stage of the network, class-specific residual feature maps are learned using the nested residual learning strategy. Furthermore, the residual feature maps added to the overall blurry image are estimated. The authors introduced a confidence parameter for each class to measure its importance. This parameter is predicted using a separate confidence network and is added to the loss function to reweight the contribution of the loss of each class to the total loss. In this way, the trained network can offset the imbalance problem in the estimation of different classes, enabling it to better recover the details of small areas such as the nose and eyes.

When the semantic segmentation maps of faces are used, the accuracy of the segmentation can significantly affect the restoration performance. Moreover, it is very difficult to obtain accurate semantic segmentation maps from blurry input images. Zhang et al. \cite{zhang2021face} employed focal loss \cite{lin2017focal} to fine-tune a face parsing network to acquire a more accurate facial structure from a blurry image. They also proposed a separate normalization and adaptive denormalization block and a texture extractor to enhance the texture and facial details of the deblurred image. Lee et al. \cite{leeProgressiveSemanticFace2020} proposed a method for learning facial component restoration without performing any segmentation. The proposed multi-semantic progressive learning (MSPL) framework was based entirely on the GAN structure. It introduces semantic coarse-to-fine progressive learning into a face deblurring task for the first time. In general, for image deblurring, progressive learning is mainly based on multi-scale network frameworks. In this model, the generator gradually generates face components in the order of skin, hair, interior (eyes, nose, and mouth), and, finally, the entire face. The multi-semantic discriminator of the model processes the multiple outputs of the generator, which allows the recovery of more realistic facial components at all intermediate layers.

In addition to face parsing maps, Lin et al. \cite{linLearningDeblurFace2020} used face sketches as priors to guide the blur-kernel estimation process. Sketches of human faces can be used to model global relationships, which can be used as prior information to achieve better results than extracting local sharp edges. The authors used a CNN to generate face sketches and used the learned sketches to guide the motion blur estimation module. Moreover, the generated sketches are also fed into the de-artifact subnetwork to encourage image reconstructions that preserve details and edges.

\begin{figure*}[h!t]
	\centering
	\includegraphics[width=0.75\linewidth]{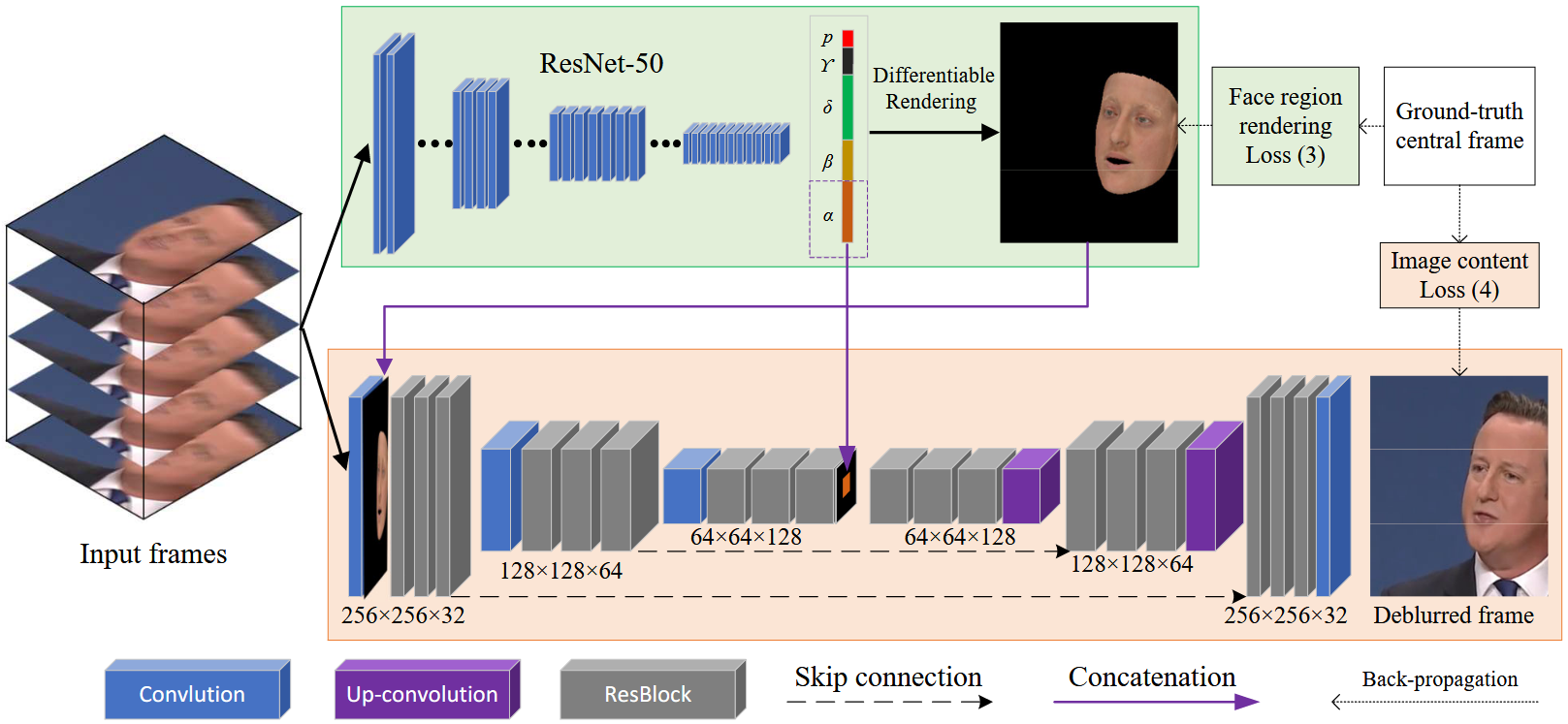}
	\caption{Deblurring neural network using a 3D human face as a prior \cite{renFaceVideoDeblurring2019}. The top green block shows how to reconstruct a 3D face by regressing the 3D morphable model \cite{paysan20093d} coefficients and render a sharp facial image. The bottom orange block focuses on face deblurring guided by the extracted identity vector and the rendered sharp face structure from the 3D face reconstruction branch.}
	\label{fig8}
\end{figure*}

Ren et al. \cite{renFaceVideoDeblurring2019} used 3D face priors \cite{hu2021face} as the guidance information to achieve facial video deblurring. Their network model is illustrated in Fig. ~\ref{fig8}. Textured 3D faces are first generated for the center frame of the video using a 3D face reconstruction network that provides image-level (e.g., intensity with sharp edges) and perceptual-level (e.g., identity) information. The face deblurring network then applies the rendered, pose-aligned facial image as a guide to recover sharp faces. Furthermore, to encourage the generation of identity-related facial details during the deblurring process, the network further embeds the identity vectors extracted by the 3D reconstruction network into the deblurring network branch. Benefiting from the clear face structure estimated by 3D face rendering, the model can achieve good output results without using a multi-scale structure and certain tricks such as perceptual loss.

\begin{figure}[h!t]
	\centering
	\includegraphics[width=1\linewidth]{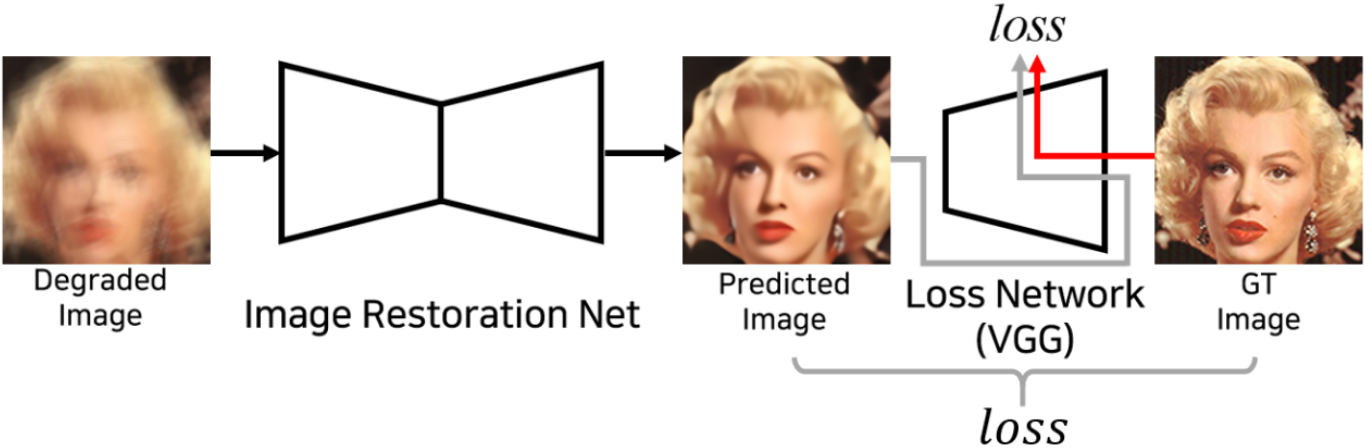}
	\caption{Deblurring neural network using deep features as the network loss \cite{jungDeepFeaturePrior2022}. ``GT'' indicates ``ground truth''.}
	\label{fig9}
\end{figure}

\begin{figure*}[h!t]
	\centering
	\includegraphics[width=0.65\linewidth]{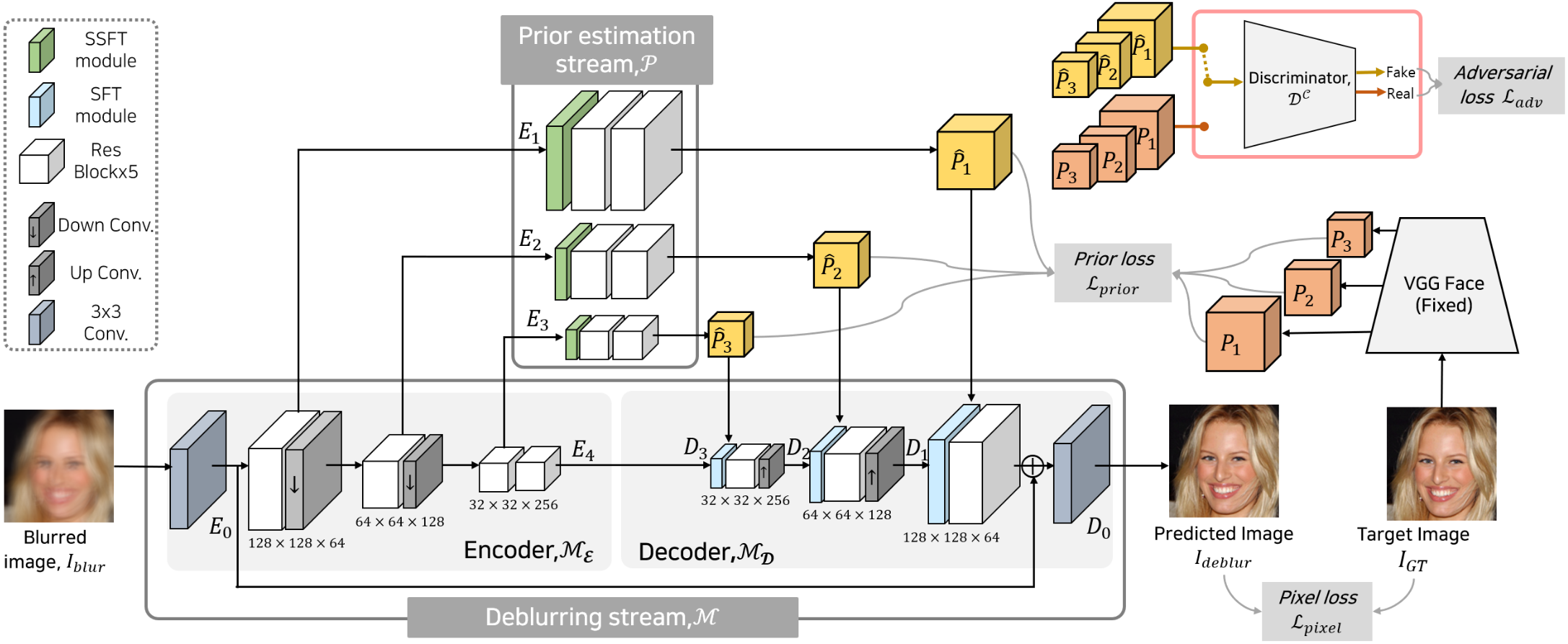}
	\caption{Deblurring neural network using the deep features of images as a network prior \cite{jungDeepFeaturePrior2022}. The deep features are extracted by the prior estimation stream. ``SFT'' indicates spatial feature transform module and ``SSFT'' indicates self-spatial feature transform module.}
	\label{fig10}
\end{figure*}

In 2017, Xu et al. \cite{xu2017learning} proposed the introduction of perceptual loss into a face deblurring network to guide the training of the network. The perceptual loss calculates the error between the deep features of two images. Its framework is shown in Fig. ~\ref{fig9}. Unlike Xu, Jung et al.  \cite{jungDeepFeaturePrior2022} tried to guide the deblurring process by using the deep features of the image as a separate stream in the network. Its network architecture is shown in Fig. ~\ref{fig10}. Both face geometry and texture information are included in the deep features. In addition, a channel attention feature discriminator was proposed to assign different importance levels to different channels of the deep feature, thereby enabling the generator to focus on more important channels. 

In 2021, Wang et al. \cite{wang2021towards} leveraged diverse generative facial priors for blind face restoration. This method also uses the deep features of images as prior information; the difference is that it uses StyleGAN2 \cite{karras2020analyzing} to extract the deep features. The input image is mapped to the closest latent code using a multilayer perceptron. The latent code is inputted into the StyleGAN2 network, and its intermediate features are extracted as generative prior information, which are modulated to guide the network to obtain more realistic results. Zhu et al. \cite{zhu2022blind} proposed an adaptive feature fusion block to synthesize the shape feature priors and deep feature priors of images. The shape feature prior of the image is acquired using a 3D face reconstruction module, and the acquisition of the deep feature prior is similar to that in \cite{wang2021towards}, which is acquired using StyleGAN2. However, these two methods focus on the comprehensive degradation of facial images including blur, noise, and JPEG compression. However, they are not designed specifically for deblurring.

In conclusion, end-to-end deblurring networks based on deep learning outperform traditional methods that require iterative optimization in terms of speed. Properly introducing semantic information into the network will improve the deblurring accuracy of the network. A common disadvantage of these methods is that they are only suitable for processing uniform motion blur and aligned facial images. These methods may fail for nonuniform blur or face profile images.

\subsection{Unsupervised learning}
On the one hand, the performance of deep neural networks strongly depends on a large number of paired training datasets; however, it is difficult to construct such paired training datasets. On the other hand, most of the current methods operate on synthetic training datasets. The learned network does not generalize well to real-world blurry images because of the large differences in the data distribution between different domains. To overcome these difficulties and limitations, some researchers focus on unsupervised methods for facial image deblurring. 

\begin{figure}[h!t]
	\centering
	\includegraphics[width=1\linewidth]{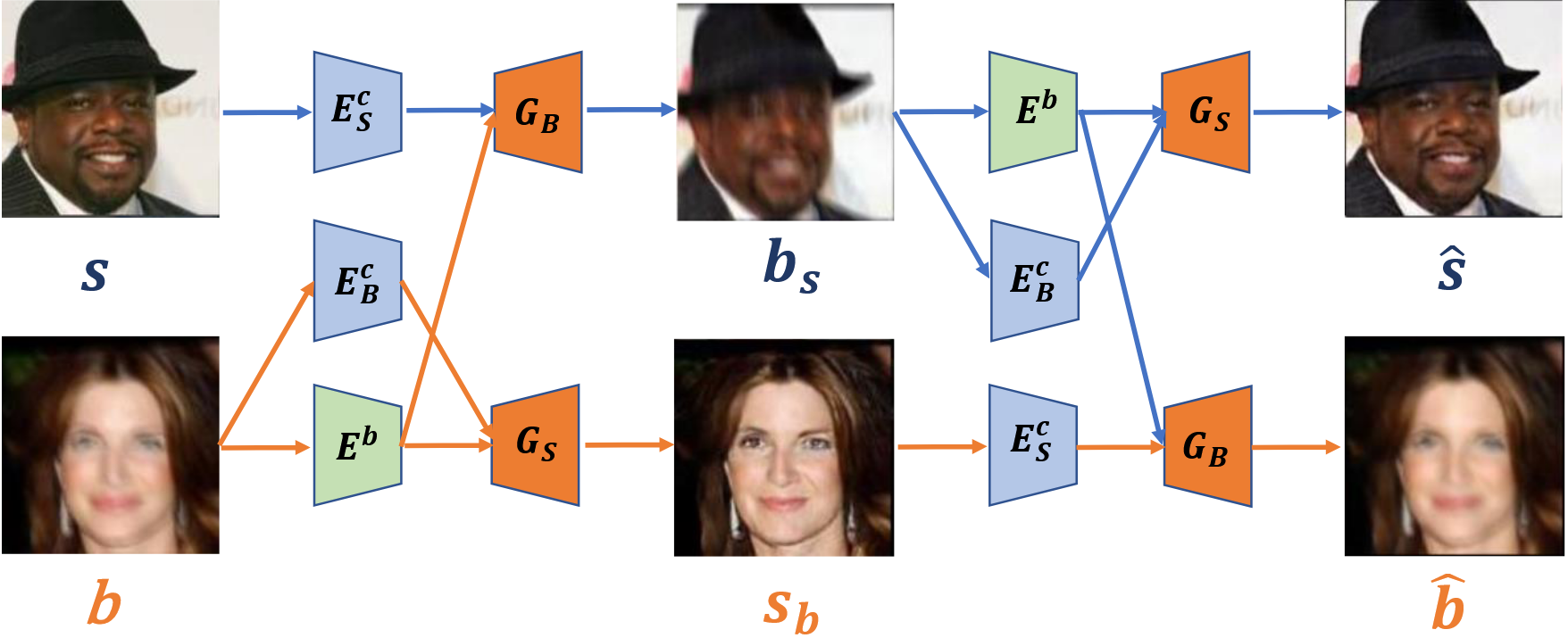}
	\caption{Unpaired methods for facial image deblurring \cite{luUnsupervisedDomainSpecificDeblurring2019}. They entirely adopted CycleGAN, including the blurring branch (top row) and deblurring branch (bottom row). $E^c_B$ and $E^c_S$ are the content encoders for the blurry and sharp images, respectively; $E^b$ is the blur encoder; and $G_B$ and $G_S$ are the blurry and sharp image generators, respectively. }
	\label{fig11}
\end{figure}

In 2018, Madam et al. \cite{madamUnsupervisedClassSpecificDeblurring2018} proposed a GAN-based method for unsupervised image deblurring. They added a reblur loss and a multi-scale gradient loss to the model. Although they achieved good performance on synthetic datasets, their results for some real blurry images were not satisfactory. In 2019, Lu et al. \cite{luUnsupervisedDomainSpecificDeblurring2019} proposed an unsupervised facial image deblurring method based on disentangled representation. The model separates the content features and blur features from the blur images and normalizes the distribution range of the extracted blur attributes by enforcing the Kullback-Leibler (KL) divergence loss, thereby realizing the decoupling of the two features. Like in CycleGAN \cite{zhu2017unpaired}, an adversarial loss and a cycle-consistency loss are used as regularizers to help the generator network generate more realistic images while preserving the content of the original images. A perceptual loss is also added to remove artifacts in blurry images. Its model architecture is shown in Fig.~\ref{fig11}. 

In the same year, Xia and Chakrabarti \cite{xia2019training} proposed a data-based unsupervised framework for training image estimation networks. This framework can be applied to a general class of observation models, where measurements are linear functions of real images accompanied by additive noise. The authors provided solutions for blind restoration and nonblind restoration, which can be used for face deblurring. In blind image restoration, a parameter estimation network and an image estimator are trained separately, and the network is guided to learn image deblurring from unlabeled data by defining a ``swap-measurement'' loss and a ``self-measurement'' loss. The training of this framework does not require paired training data, but requires two blurry images with different blur types for the same scene. 

Unsupervised learning methods lack corresponding ground-truth images as supervision information; therefore, the generated images are of low quality. However, these methods allow training on easily generated unpaired data as well as on wild images that represent real blur; therefore, it is of great significance to develop unsupervised learning methods.

\begin{table*}[h!t]
		\center
		\caption{Overview of the single facial image deblurring methods.}
		\begin{tabular}{c|c|ccc}
			\toprule
			Method&Category&Architecture&Loss&Type of Prior \\
			\midrule
			Hacohen et al. \cite{hacohenDeblurringExampleUsing2013} &	Model-based&Iterative optimization&None&Reference image\\
			Pan et al. \cite{pan2014deblurring}&Model-based&Iterative optimization&None&Reference image\\
			Anwar et al. \cite{anwar2018image}&Model-based&Iterative optimization&None&Class-specific prior\\
			\hline

			Madam et al. \cite{madamUnsupervisedClassSpecificDeblurring2018}&Unsupervised&GAN, CNN&\begin{tabular}{c}Adversarial loss;\\Reblurring loss;\\Gradient loss\end{tabular}&None\\
			\cline{3-5}
			Lu et al. \cite{luUnsupervisedDomainSpecificDeblurring2019}&Unsupervised&GAN&\begin{tabular}{c}Adversarial loss;\\Cycle-consistency loss;\\Perceptual loss;\\KL divergence loss\end{tabular}&None	\\
			\cline{3-5}
			Xia and Chakrabarti \cite{xia2019training}&Unsupervised&Arbitrary&\begin{tabular}{c}Swap-measurement loss;\\
				Self-measurement loss\end{tabular}&None\\ 	
			\hline

			Chrysos et al. \cite{chrysosDeepFaceDeblurring2017}&Supervised&ResNet&Huber loss \cite{huber1973robust}&Facial landmarks\\
			\cline{3-5}
			Song et al. \cite{song2019joint}&Supervised&CNN&L2 content loss&Facial masks\\
			\cline{3-5}
			Shen et al. \cite{shenDeepSemanticFace2018}&Supervised&GAN, Multi-scale&\begin{tabular}{c}L1 content loss;\\Local structural loss;\\Perceptual loss;\\Adversarial loss\end{tabular}&Face semantic labels\\
			\cline{3-5}
			Jin et al. \cite{jinLearningFaceDeblurring2018}&
			Supervised&ResNet&Content loss&Large receptive field\\ 
			\cline{3-5}
			Ren et al. \cite{renFaceVideoDeblurring2019}&Supervised&ResNet, U-Net&\begin{tabular}{c}L2 content loss;\\Rendering loss\end{tabular}&Rendered 3D face\\
			\cline{3-5} 
			Chrysos et al. \cite{chrysosMotionDeblurringFaces2019}&Supervised&GAN, U-Net&\begin{tabular}{c}L1 content loss;\\
				Projection loss;\\
				Adversarial loss\end{tabular}&None\\ 
			\cline{3-5}
			Lin et al. \cite{linLearningDeblurFace2020}&Supervised&GAN&\begin{tabular}{c}L1 content loss;\\Adversarial loss\end{tabular}&Face sketches\\  
			\cline{3-5}
			UMSN \cite{yasarla2020deblurring}&Supervised&ResNet&\begin{tabular}{c}Local structural loss;\\	Perceptual loss\end{tabular}&Face semantic labels\\ 
			\cline{3-5}
			Shen et al. \cite{shenExploitingSemanticsFace2020}&Supervised&\begin{tabular}{c}GAN, U-Net,\\Multi-scale\end{tabular}&\begin{tabular}{c}Cross-entropy loss;\\L1 content loss;\\Local structural loss;\\Perceptual loss;\\Adversarial loss\end{tabular}&Face semantic labels\\ 
			\cline{3-5}
			MSPL \cite{leeProgressiveSemanticFace2020}&Supervised& GAN &\begin{tabular}{c}Local structural loss;\\Perceptual loss;\\Adversarial loss\end{tabular}&Face semantic labels\\
			\cline{3-5}
			GFP-GAN \cite{wang2021towards}&Supervised& U-Net, GAN&\begin{tabular}{c}L1 content loss;\\Perceptual loss;\\Adversarial loss;\\Local structural loss;\\Identity preserving loss\end{tabular}&Generative prior in StyleGAN\\
			\cline{3-5} 
			Qi et al. \cite{qiBlindFaceImages2021}&Supervised&U-Net, GAN&\begin{tabular}{c}L2 content loss;\\Perceptual loss;\\Adversarial loss\end{tabular}&None\\ 
			\cline{3-5}
			DFPGnet \cite{jungDeepFeaturePrior2022}&Supervised&U-Net, GAN & \begin{tabular}{c}L1 content loss;\\
				Adversarial loss;\\Prior loss\end{tabular}&Deep feature prior in VGGFace\\
			\cline{3-5}
			SGPN \cite{zhu2022blind}&Supervised& GAN & \begin{tabular}{c}L1 content loss;\\
				Adversarial loss\end{tabular}&\tabincell{c}{Generative prior and\\ rendered 3D face prior}\\
			\bottomrule
		\end{tabular}
		\label{tab4}
	\end{table*}

\section{Datasets used for facial image deblurring}
Depending on how the datasets are constructed, they can be divided into real-shot datasets and synthetic datasets. For real-shot datasets, blurry images are usually obtained by averaging the video frames or moving the camera through a specific trajectory. For facial image deblurring, most methods are trained on synthetic blurry-sharp image pairs.

\subsection{Real-shot datasets}
\textbf{$2MF^2$ dataset} \cite{chrysosDeepFaceDeblurring2017}: This dataset was created by processing and extracting faces from videos containing facial images. It consists of 1150 videos containing 2.1 million frames of acceptable facial images, and landmark localization is implemented for each image. The authors later updated the dataset \cite{chrysosMotionDeblurringFaces2019}, expanding the number of videos and face identities in the images. The augmented dataset contains people of different ages and ethnicities and includes 19 million frames from 11,590 videos of 850 different identities, each of which appears in multiple videos. Moreover, the authors generated motion-blurred images by averaging multiple frames of the same face. 

\textbf{Lai et al. dataset} \cite{laiComparativeStudySingle2016}: Lai et al. collected real-world blurry images obtained in the wild. These images were captured by different users using different cameras and settings. There are 100 real blurry images, including natural, facial, and text images, none of which has any corresponding ground truth. Various face deblurring models and methods were tested and compared on this dataset to evaluate their performance on real blurry images.

\subsection{Synthetic datasets}
\textbf{Shen et al. dataset} \cite{shenDeepSemanticFace2018}: Shen et al. collected 2000, 2164, and 2300 clear facial images as ground truth from the Helen dataset \cite{leInteractiveFacialFeature2012}, CMU PIE dataset \cite{terencesimCMUPoseIllumination2003}, CelebA dataset \cite{liu2015deep}, respectively. Then, 20,000 motion blur kernels were generated according to the method in \cite{boracchiModelingPerformanceImage2012}, with sizes ranging from 13$\times$13 to 27$\times$27. The corresponding blurry images were synthesized by convolving clear images with these blur kernels and adding Gaussian noise. In total, 130 million blurry-sharp image training pairs were generated. The authors also constructed a test set that contained 16,000 images. It was generated by collecting 100 facial images from the validation set of the Helen and CelebA datasets and convolving them with another 80 generated motion blur kernels.

\textbf{MSPL dataset} \cite{leeProgressiveSemanticFace2020}: Lee et al. used 30,000 HR facial images from the CelebA-HQ dataset \cite{leeMaskGANDiverseInteractive2020} as clear ground truth and generated 18,000 motion blur kernels according to the method in \cite{chakrabarti2016neural}. They convolved the sharp image with a kernel and added Gaussian noise. The generated images were split into two subsets: 24,183 image pairs for training and 5817 image pairs for validation. The authors also collected 240 clear images (80 in each dataset) from the CelebA , CelebA-HQ, and FFHQ datasets \cite{karras2019style} and convolved them with 240 synthetic motion blur kernels for model testing. Two types of test data were generated: MSPL-Center and MSPL-Random. MSPL-Random was generated by random rotation, cropping, and horizontal flipping of MSPL-Center. 

\textbf{Jin et al. dataset} \cite{jinLearningFaceDeblurring2018}: Jin et al. collected and cropped 110,000 clear facial images from FaceScrub \cite{ng2014data} and generated 10K random motion blur kernels according to the method in \cite{chakrabarti2016neural}. Finally, white Gaussian noise was added to generate the training data. 

\textbf{Lin et al. dataset} \cite{linLearningDeblurFace2020}: Lin et al. collected 2184, 2000, 2000, and 2400 clear facial images from the CMU PIE, Helen, CelebA, and PubFig datasets \cite{kumarAttributeSimileClassifiers2009}, respectively. These clear images were convolved with 20,000 motion blur kernels to generate blurry images. The blur kernels varied in size from 21$\times$21 to 51$\times$51. To increase the diversity of the data, the authors applied data augmentation, including random rotation, cropping, and scaling.

Many studies chose different datasets and blur kernel generation methods to simulate the training or testing of their own image pairs according to their specific application scenarios; these datasets are summarized in Table \ref{tab1}. Some of them are not available to the public and are used only for training their models.
\begin{table*}[h!t]
	\center
	\caption{Overview of the different datasets constructed by different authors for use in training or testing.}
	\begin{tabular}{ccccc}
		\toprule
		Dataset & Synthetic & Real & Clear image source & Blur kernel source \\
		\midrule
		$2MF^2$  \cite{chrysosDeepFaceDeblurring2017} & \ding{56} & \ding{52}   & None & None  \\
		Lai et al. \cite{laiComparativeStudySingle2016} & \ding{56} & \ding{52}   & None & None \\
		Shen et al. \cite{shenDeepSemanticFace2018} & \ding{52} &\ding{56}   & Helen, CMU PIE, and CelebA & \cite{boracchiModelingPerformanceImage2012}  \\
		MSPL \cite{leeProgressiveSemanticFace2020} & \ding{52} &\ding{56} &CelebA-HQ   & \cite{chakrabarti2016neural} \\
		Jin et al. \cite{jinLearningFaceDeblurring2018} & \ding{52} &\ding{56}  & FaceScrub & \cite{chakrabarti2016neural} \\
		Lin et al. \cite{linLearningDeblurFace2020} & \ding{52} &\ding{56}   & CMU PIE, Helen, CelebA, and PubFig &
		\cite{boracchiModelingPerformanceImage2012} \\
		Madam \cite{madamUnsupervisedClassSpecificDeblurring2018}, Lu \cite{luUnsupervisedDomainSpecificDeblurring2019}, and Qi \cite{qiBlindFaceImages2021}& \ding{52} & \ding{56}   & CelebA & \cite{chakrabarti2016neural}  \\
		Wang et al. \cite{wang2021towards} & \ding{52} & \ding{56}   & FFHQ & Gaussian blur kernel  \\
		Xia \cite{xia2019training} and Yasarla \cite{yasarla2020deblurring} & \ding{52} & \ding{56}   & Helen and CelebA & \cite{chakrabarti2016neural}  \\
		Song et al. \cite{song2019joint} & \ding{52} & \ding{56}   & Multi-PIE \cite{gross2010multi} and PubFig & Gaussian blur kernel  \\
		\bottomrule
	\end{tabular}
	\label{tab1}
\end{table*}

\section{Training strategy}

\subsection{Loss function}
When a network is trained, the choice of loss function is critical to the final deblurring performance. In general deblurring models, content loss, perceptual loss, and adversarial loss are often combined to achieve better results. For domain-specific face deblurring, local structure loss, identity preserving loss, etc., have also been proposed to help improve the performance.

\textbf{Content loss. }Content loss, also known as reconstruction loss, is the most classic and commonly used loss function. Its goal is to measure the pixel-by-pixel difference between the output deblurred images and the ground-truth images. $L_1$ distance (mean absolute error) or $L_2$ distance (mean squared error) is often used to measure the difference. Its calculation formula is as follows:
\begin{equation}
L_{\text {content }}=\frac{1}{W H} \sum_{i=1}^{W} \sum_{j=1}^{H} f\left(I_{D}(i, j)-I_{S}(i, j)\right)
\end{equation}

where the function $f$ represents the $L_1$ distance or the $L_2$ distance; $I_D$ and $I_S$ represent the deblurred image and the corresponding sharp ground-truth image, respectively; and $W$ and $H$ represent the width and height of the image, respectively. As can be seen from the formula, the content loss calculates the distance of each pixel in the two images and then sums it up. When content loss is used in the network, the pixel values of the deblurred output image should be as close to the ground truth as possible.

However, this method of comparing differences pixel by pixel is relatively simple and crude, and does not consider human subjective visual perception. Adopting content loss tends to lead to oversmoothed output results. Some methods \cite{chrysosMotionDeblurringFaces2019,madamUnsupervisedClassSpecificDeblurring2018} also incorporate a first-order gradient distance between the deblurred image and the ground-truth image in the content loss to remove undesired ringing artifacts at image boundaries.

\textbf{Perceptual loss. }A visually pleasing result cannot be achieved using content loss alone. Perceptual loss has been widely used in applications, such as style transfer and image super-resolution, to enhance the perceptual effect. It aims to compare the distance between the deblurred and sharp images in a high-dimensional feature space to measure their feature similarity. Perceptual loss can be written as
\begin{equation}
L_{p e r}=\left\|\phi_{l}\left(I_{D}\right)-\phi_{l}\left(I_{s}\right)\right\|_{2}^{2}
\end{equation}

where $\phi_{l}$ represents the output of layer $l$ of a pretrained network. The authors in \cite{shenExploitingSemanticsFace2020,song2019joint,yasarla2020deblurring,leeProgressiveSemanticFace2020,linLearningDeblurFace2020} used the outputs of the $pool2$ and $pool5$ layers of the pretrained VGGFace network, whereas those in \cite{qiBlindFaceImages2021,wang2021towards,jungDeepFeaturePrior2022} used the outputs of $conv3\_3$ and other layers of the pretrained VGG-19 network as features. The low-level features in the pretrained network contain edge, color, brightness, and other information, whereas the high-level features contain texture and rich semantic information. When perceptual loss is used in the network, the details and texture of the output image can be enhanced. For example, the edge parts of the image will be sharper, which compensates for the over-smoothed output results caused by content loss.  

\textbf{Adversarial loss. }To make the output facial images as realistic as and as close to natural images as possible, researchers have developed various GAN-based methods. GAN-based networks define the problem as a min-max optimization process to make the output image of the generator look closer to the real face. The training objectives of generator $G$ and discriminator $D$ are opposite, and they generate real images by competing against each other. Adversarial loss is calculated as follows:
\begin{equation}
\left.L_{a d v}=E\left[\log D\left(I_{S}\right)\right]+E\left[\log \left(1-D\left(G\left(I_{B}\right)\right)\right)\right)\right]
\end{equation}

where $I_{B}$ represents the blurry facial image. Through the competition between the generator and the discriminator, the adversarial loss will be able to make the output image of the network look as close to the real face as possible. However, training with only adversarial loss may result in the loss of image details (such as the eyes, nose, and mouth of faces). This is because the generator can generate real images even if some details or colors are lost, and the discriminator classifies these images as real with a small adversarial loss.

\textbf{Local structural loss.} Content loss supervises the image as a whole owing to the small proportion of the eyes, nose, mouth, and other areas of the face in the entire image; using content loss solely cannot restore these parts well. Many methods \cite{shenDeepSemanticFace2018,shenExploitingSemanticsFace2020,yasarla2020deblurring,wang2021towards} introduced a local structure loss of the face to enhance important facial components in terms of visual perception. A common practice is to extract the face parsing map of the input image or directly perform face segmentation and then apply different importance weights to different facial components. Local structural loss is defined as follows:
\begin{equation}
L_{s}=\sum_{i=1}^{N} \omega_{i} \cdot\left\|M_{i} \odot I_{D}-M_{i} \odot I_{s}\right\|
\end{equation}

where $M_i$ is the semantic mask of the $i$-th component extracted from the facial image and $\omega_{i}$ refers to the weight of the $i$-th component. The models typically apply local structural loss to components such as the eyes, eyebrows, teeth, lips, and nose, but not to areas such as the skin and hair. The introduction of local structure loss can force the network to reconstruct sharper details in the key component parts of the face.

\textbf{Identity preserving loss. }Similar to perceptual loss, identity preserving loss limits the distance between the deep features of the two images. In \cite{wang2021towards}, the authors adopted the facial recognition network ArcFace\cite{deng2019arcface} to extract the identity features of facial images. By limiting the distance between the identity feature of the deblurred image and the real sharp image, the network prevents the distortion of the deblurring result, thereby improving the accuracy of the subsequent face recognition and other tasks. 

\textbf{Other losses.} In addition to the aforementioned losses, some specific loss functions exist to improve the performance of specific models. Unsupervised methods \cite{madamUnsupervisedClassSpecificDeblurring2018} define reblurring loss to address the lack of ground-truth images. Another unsupervised method proposed in \cite{xia2019training} addresses the lack of training image pairs by defining ``self-measurement'' loss and ``swap-measurement'' loss. A prior loss was introduced in \cite{jungDeepFeaturePrior2022} to ensure that the deep feature priors used to guide the network training are accurate. To extract accurate 3D face priors, Ren et al. \cite{renFaceVideoDeblurring2019} defined a rendering loss function \ to ensure that the 3D face reconstruction process is unaffected by blurry images.

 Each loss function has different effects on the deblurred images. In the actual training process, it is necessary to comprehensively select a variety of loss functions according to the actual needs of the network model to improve network performance.

\subsection{Training scheme}
A face deblurring model may contain multiple subnetworks, and these subnetworks or the entire network is trained separately with various loss functions. Moreover, the training strategy of the large network will also affect the final result to a certain extent. A progressive training strategy was adopted by \cite{shenExploitingSemanticsFace2020}. First, they trained a subnetwork. The training scheme is as follows: with the remaining networks fixed, each subnetwork is trained separately with a specific partial loss function as a coarse adjustment. Then, all subnetworks, i.e., the entire model, are jointly trained by minimizing the overall loss. 

In addition, to improve the model performance and enable it to handle random blur kernels in real-world blurry images, Shen et al. \cite{shenDeepSemanticFace2018} proposed an incremental learning strategy. The authors synthesized a large number of blurry-sharp training pairs consisting of different blur kernel sizes. Direct training on a large amount of data can easily lead to a model falling into a locally optimal solution. The incremental learning strategy gradually increases the training data and the motion blur kernel size during the training process. The network is first trained on small blur kernels, and then the training set is progressively expanded to add large blur kernels. Then, the network is trained on the set of original and new blur kernels together. Larger blur kernels are continuously added until the training data include all the blur kernels.

\section{Evaluation metrics}
Evaluating the output image quality can help in judging the quality of a model. However, it is difficult to construct a sufficiently objective  metric that conforms to human perception. In the field of face deblurring, the metrics often used for comparison can be divided into three categories, according to different levels and purposes. The first type of metric aims to provide an evaluation from the pixel level of the images. The second type evaluates the quality of the images in terms of visual perception by extracting the deep features of the images. The third type is task oriented, which evaluates the quality of images by comparing their accuracy in advanced visual tasks.

\subsection{Image-level evaluation metrics}
Among these metrics, the most commonly used are peak signal-to-noise ratio (PSNR) and structural similarity (SSIM). These metrics do not require any additional inputs and can be obtained through simple calculations. When calculating, they require the ground truth corresponding to the images, which are the reference evaluation metrics.

\textbf{PSNR:} This metric is acquired by calculating the pixel-level mean squared error of two images. The larger the value, the smaller the difference between the two images. However, the numerical results of PSNR are often inconsistent with the subjective perception of human vision. Using the PSNR metric tends to lead to oversmoothed results.

\textbf{SSIM} \cite{wangImageQualityAssessment2004}: SSIM is modeled after the visual system of humans. This metric measures the difference between two images in terms of brightness, contrast, and structure. The mean, variance, and covariance of the images are used to evaluate their brightness, contrast, and structural similarity, respectively. The larger the value of SSIM, the more similar the two images. However, this metric is not a good representation of how humans actually feel. Images with lower SSIM values may also have a good visual experience.

\subsection{Perceptual evaluation metrics}
These metrics mine deep features in images and can often reflect results that are consistent with human vision. They include reference evaluation metrics that require a ground truth and no-reference evaluation metrics that do not require clear images.

\textbf{LPIPS} \cite{zhangUnreasonableEffectivenessDeep2018}: Different from the above metrics, LPIPS calculates the distance between two images in the high-dimensional feature space to make the calculation result as close as possible to human visual perception. High-dimensional deep features are extracted using a pretrained classification network. LPIPS is a perceptual metric, and a smaller value indicates that two images are visually similar. The calculation formula is as follows:
\begin{equation}
d\left(x, x_{0}\right)=\sum_{l} \frac{1}{H_{l} W_{l}} \sum_{h, w}\left\|w_{l} \odot\left(\hat{y}_{h w}^{l}-\hat{y}_{0 h w}^{l}\right)\right\|_{2}^{2}
\end{equation}
where $d$ is the distance between $x$ and $x_0$, $\hat{y}_{h w}^{l}$ represents the feature extracted from the network, and $w_{l}$ is used to scale channel $l$. Finally, the $L_2$ distance is calculated, the average in space is taken, and the channel is summed up. 

\textbf{NIQE} \cite{mittalMakingCompletelyBlind2013}: This metric constructs a set of quality perception features and fits them to a multivariate Gaussian model. The characteristics of quality perception are extracted using the natural scene statistics (NSS) model. Then, the quality of the test image is given as the distance between the multivariate Gauss (MVG) fitting of the NSS features extracted from the test image and the MVG model of the quality perception features extracted from the natural image corpus. The calculation formula is as follows:
\begin{equation}
	\begin{split}
	&D\left(\nu_{1}, \nu_{2}, \Sigma_{1}, \Sigma_{2}\right)=\\
&\sqrt{\left(\left(\nu_{1}-\nu_{2}\right)^{T}\left(\frac{\Sigma_{1}+\Sigma_{2}}{2}\right)^{-1}\left(\nu_{1}-\nu_{2}\right)\right)}
	\end{split}
\end{equation}
where $\nu_{1}$ and $\nu_{2}$ are the mean vectors and $\Sigma_{1}$ and $\Sigma_{2}$ are the covariance matrices of the natural MVG model and distorted image MVG model. The smaller the value, the better the image quality.

In addition, there are many no-reference evaluation metrics for the evaluation of deblurred facial images, such as PI \cite{ignatovPIRMChallengePerceptual2019}, BRISQUE \cite{mittalNoReferenceImageQuality2012}, NRQM \cite{ma2017learning}, PIQE \cite{nBlindImageQuality2015}, and FID \cite{heuselGANsTrainedTwo2017}.

\subsection{Advanced visual task evaluation metrics}

Image deblurring is a low-level task that aims to improve the accuracy of higher-level tasks, such as image recognition, object detection, and image segmentation. Therefore, for the face deblurring task, the meaningful metrics are the identity invariance of the deblurred faces and the accuracy of the face recognition. 

\textbf{Identity distance.} To measure the face identity similarity between deblurred images and ground-truth images, many methods compute their feature distance $d_{VGG}$ in pre-trained face recognition networks such as VGGFace \cite{parkhi2015deep}. The smaller the feature distance, the more similar the face identities in the two images. 

\textbf{Face detection} \cite{shenDeepSemanticFace2018,shenExploitingSemanticsFace2020,yasarla2020deblurring,jungDeepFeaturePrior2022}. For detection testing, the OpenFace \cite{amos2016openface} toolbox is used to measure the success rate of face detection in deblurred images. 

\textbf{Face recognition} \cite{chrysosMotionDeblurringFaces2019,shenDeepSemanticFace2018,shenExploitingSemanticsFace2020,jungDeepFeaturePrior2022}. For face verification, networks such as MobileNet \cite{howard2017mobilenets} trained with ArcFace \cite{deng2019arcface} loss are used to compare the face recognition accuracies \cite{huang2008labeled} of the output deblurred images of different methods.

\section{Performance evaluation}
Because model-based methods do not require training datasets, they cannot be compared directly with learning-based methods. Specifically, model-based methods assist in the optimization process by designing a specific prior as a regularization term or building a sample library. However, learning-based methods often need to train the network on specific datasets to ensure the effect on the test images of the same distribution. Therefore, in this section, we compare the experimental performance of representative model-based and deep learning-based methods.

\begin{table*}[h!t]
	\center
	\caption{Evaluation results of the performance of the model-based representative methods on the CMU PIE dataset. The best results are highlighted in bold.}
	\begin{tabular}{cccc}
		\toprule
		Method & Type & PSNR($\uparrow$) & SSIM($\uparrow$) \\
		\midrule
		Shan et al. \cite{shan2008high} & Sparse prior   & 25.59 & 0.775 \\
		Krishnan et al. \cite{krishnan2011blind} & Sparse prior   & 23.54 & 0.693 \\
		Cho and Lee \cite{cho2009fast} & Edge selection   & 24.38 & 0.699 \\
		Xu and Jia \cite{xu2010two} & Edge selection   & 23.30 & 0.739 \\
		Anwar et al. \cite{anwar2018image} & Class-specific prior & \textbf{30.75} & \textbf{0.881}\\
		\bottomrule
	\end{tabular}
	\label{tab2}
\end{table*}

\begin{figure*}[htbp]
	\centering
	\subfigure{
		\begin{minipage}[t]{0.11\linewidth}
			\centering
			\includegraphics[width=0.75in]{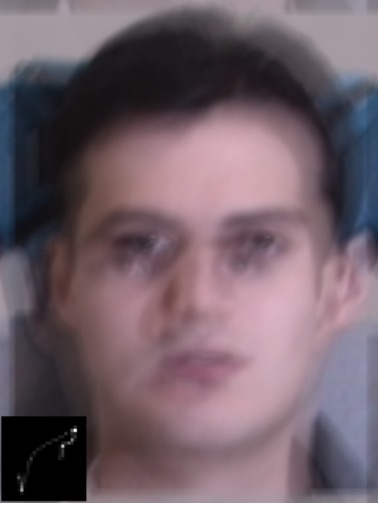}
		\end{minipage}%
	}%
	\subfigure{
		\begin{minipage}[t]{0.11\linewidth}
			\centering
			\includegraphics[width=0.75in]{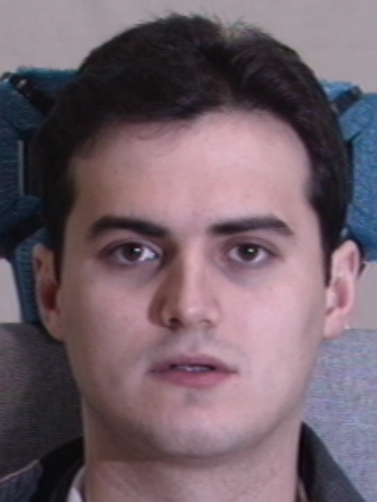}
		\end{minipage}%
	}%
	\subfigure{
		\begin{minipage}[t]{0.11\linewidth}
			\centering
			\includegraphics[width=0.75in]{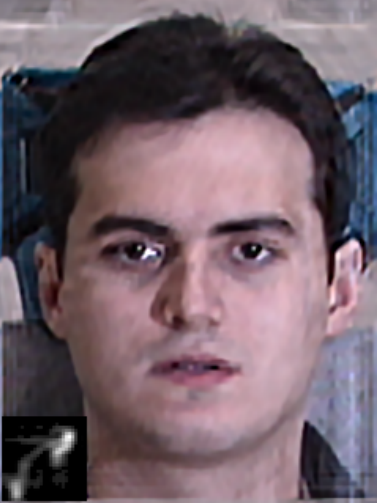}
		\end{minipage}
	}%
	\subfigure{
		\begin{minipage}[t]{0.11\linewidth}
			\centering
			\includegraphics[width=0.75in]{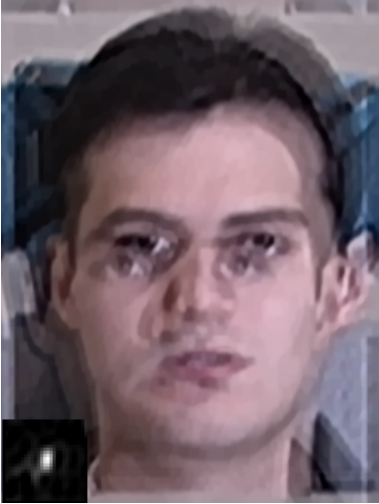}
		\end{minipage}
	}%
	\subfigure{
		\begin{minipage}[t]{0.11\linewidth}
			\centering
			\includegraphics[width=0.75in]{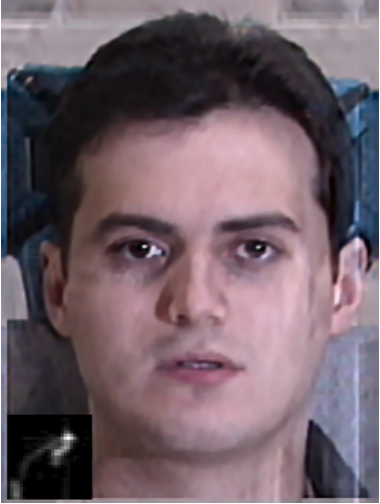}
		\end{minipage}
	}%
	\subfigure{
		\begin{minipage}[t]{0.11\linewidth}
			\centering
			\includegraphics[width=0.75in]{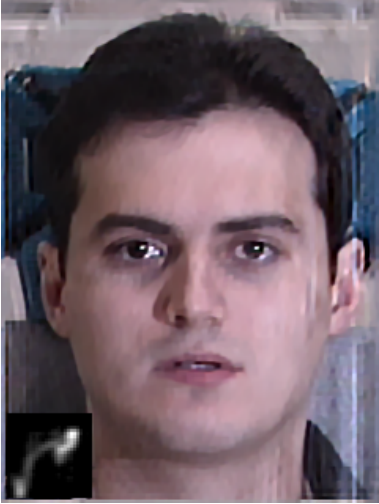}
		\end{minipage}
	}%
	\subfigure{
		\begin{minipage}[t]{0.11\linewidth}
			\centering
			\includegraphics[width=0.75in]{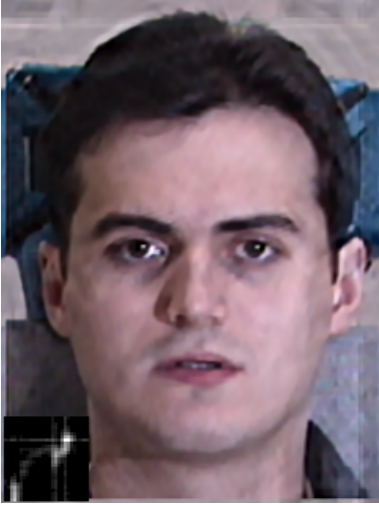}
		\end{minipage}
	}%
	\subfigure{
		\begin{minipage}[t]{0.11\linewidth}
			\centering
			\includegraphics[width=0.75in]{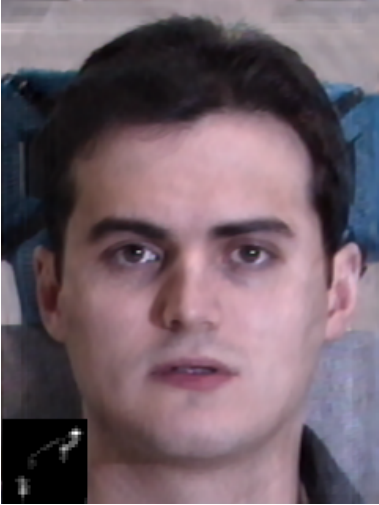}
		\end{minipage}
	}%
	\centering
	\caption{Qualitative comparisons of the representative model-based methods on blurry facial images from the CMU PIE dataset.   From left to right: blurry image, ground truth, Shan et al. \cite{shan2008high}, Krishnan et al. \cite{krishnan2011blind}, Cho and Lee \cite{cho2009fast}, Xu at al. \cite{xu2010two}, Pan et al. \cite{pan2014deblurring}, and Anwar at al. \cite{anwar2018image}.}
	\label{fig12}
\end{figure*}

\subsection{Comparison of model-based methods}
For the traditional methods, we compared four classes of methods: sparse prior-based methods \cite{shan2008high,krishnan2011blind}, class-specific prior-based methods \cite{anwar2018image}, exemplar-based methods \cite{pan2014deblurring}, and the edge selection-based general deblurring methods \cite{cho2009fast,xu2010two}.

We first show the performance of these methods on synthetic datasets, where test images were selected from the CMU PIE dataset. We chose PSNR and SSIM as the evaluation metrics,  which are given in the respective papers, and we show them in Table \ref{tab2}. Pan et al. \cite{pan2014deblurring} did not release their source code and their results for these two metrics; therefore, we only show their visual performance.  To compare the visual effects, we show their qualitative comparison examples in Fig.~\ref{fig12}, which are given in \cite{anwar2018image}, as these authors did not provide a full implementation of their methods. The bottom-left corner of the recovered image shows the blur kernel estimated using the aforementioned methods.

The methods of Shan et al. \cite{shan2008high} and Krishnan et al. \cite{krishnan2011blind} produced obvious artifacts and multiple edges (ghosting). This is because the sparse prior is a general prior and does not perform well on facial images that contain less texture information. The methods of Cho and Lee \cite{cho2009fast} and Xu and Jia \cite{xu2010two}, which were developed for general deblurring, estimate blur kernels that are denser than the ground truth. This is because there are insufficient sharp edges in the facial image to recover an accurate blur kernel. The exemplar-based method of Pan et al. \cite{pan2014deblurring} suffers from severe artifacts in the skin texture. This is because it only considers the restoration of the contour of the face without considering the correctness of the internal details of the face. The method of Anwar et al. \cite{anwar2018image} learns specific priors for different classes by learning the distribution features of different types of images in the frequency space. However, the restored images are too smooth and lose the contour information of specific parts. 

\begin{figure}[htbp]
	\centering
	\subfigure{
		\begin{minipage}[t]{0.19\linewidth}
			\centering
			\includegraphics[width=0.65in]{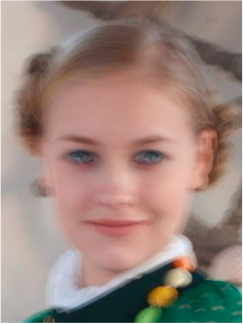}
		\end{minipage}%
	}%
	\subfigure{
		\begin{minipage}[t]{0.19\linewidth}
			\centering
			\includegraphics[width=0.65in]{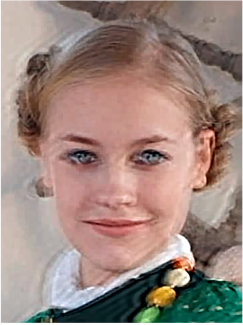}
		\end{minipage}%
	}%
	\subfigure{
		\begin{minipage}[t]{0.18\linewidth}
			\centering
			\includegraphics[width=0.65in]{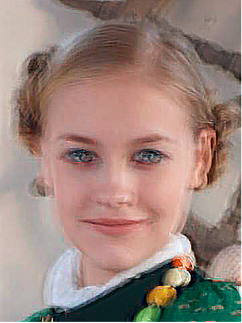}
		\end{minipage}
	}%
	\subfigure{
		\begin{minipage}[t]{0.18\linewidth}
			\centering
			\includegraphics[width=0.65in]{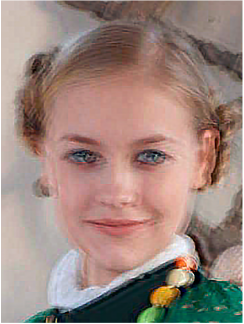}
		\end{minipage}
	}%
	\subfigure{
		\begin{minipage}[t]{0.18\linewidth}
			\centering
			\includegraphics[width=0.65in]{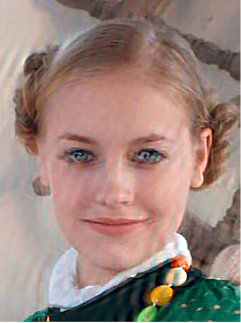}
		\end{minipage}
	}%
	\centering
	\caption{Qualitative comparisons of the representative model-based methods on real blurry facial images. From left to right: blurry image, Krishnan et al. \cite{krishnan2011blind}, Xu at al. \cite{xu2010two}, Pan et al. \cite{pan2014deblurring}, and Anwar at al. \cite{anwar2018image}.}
	\label{fig13}
\end{figure}

We also show the performance of the representative methods of the above four categories on real blurry images, which are shown in Fig. ~\ref{fig13}. These images do not have a corresponding sharp ground truth. These four classes of methods do not perform well and produce different degrees of artifact on real images. Furthermore, these methods do not restore the sharp results effectively in areas such as the eyes and nose.

\begin{table*}[h!t]
	\center
	\caption{Evaluation results of the performance of the representative methods on the Shen test dataset \cite{shenDeepSemanticFace2018}. The best results are highlighted in bold.}
	\begin{tabular}{cccccccccc}
		\toprule
		\multirow{2}{*}{Method} & \multirow{2}{*}{Type}&\multicolumn{4}{c}{Helen} & \multicolumn{4}{c}{CelebA} \\
		\cmidrule{3-10}
		&& PSNR($\uparrow$) & SSIM($\uparrow$)    & $d_{VGG}(\downarrow)$    & LPIPS($\downarrow$) &  PSNR($\uparrow$) & SSIM($\uparrow$) & $d_{VGG}(\downarrow)$ & LPIPS($\downarrow$) \\
		\midrule
		Shen et al. \cite{shenDeepSemanticFace2018} & Supervised   & 25.58 & 0.861 & 91.06  & 0.1527 & 24.34 & 0.860 & 117.50 & 0.1832\\
		UMSN \cite{yasarla2020deblurring} & Supervised   & \textbf{27.75}& 0.897 & 86.87  & 0.1086 &\textbf{26.62} & 0.908 & 66.33 & 0.1401\\
		Shen et al.\cite{shenExploitingSemanticsFace2020} & Supervised   & 25.91 & 0.869 & -  & - & 24.89 & 0.875 & - & -\\
		MSPL \cite{leeProgressiveSemanticFace2020} & Supervised   & 25.91 & 0.881 & 47.80  & \textbf{0.0828} & 24.91 & 0.885 & 57.54 & \textbf{0.0962}\\
		Wang et al. \cite{wang2021towards} & Supervised   & 22.30 & 0.775 & 206.57  & 0.1592 & 21.62 & 0.792 & 261.50 & 0.1503\\
		DFPGnet \cite{jungDeepFeaturePrior2022} & Supervised   & 27.70 & \textbf{0.911} & \textbf{42.84}  & 0.0928 & 26.56 & \textbf{0.915} & \textbf{53.38} & 0.1052\\
		Lu et al. \cite{luUnsupervisedDomainSpecificDeblurring2019} & Unsupervised   & 20.25 & 0.705 & 241.93  & 0.1654 & 19.96 & 0.742 & 305.96 & 0.1688\\
		\tabincell{c}{Xia and\\ Chakrabarti \cite{xia2019training}} & Unsupervised   & 26.13 & 0.886 & 55.97  & 0.1052 & 25.18 & 0.892 & 68.05 & 0.1199\\
		\bottomrule
	\end{tabular}
	\label{tab3}
\end{table*}

\subsection{Comparison of learning-based methods}
We compared the methods based on semantic segmentation maps, methods based on generative priors, blind face restoration methods, and unsupervised face deblurring methods. We compared the performance of these methods on the two most commonly used datasets: the Shen et al. dataset \cite{shenDeepSemanticFace2018} and the MSPL dataset. The Shen et al. dataset was constructed from the Helen and CelebA datasets; therefore, we compared the performance of the above methods on these datasets separately. For the evaluation metrics, PSNR and SSIM are the most commonly used metrics; however, they cannot represent the most realistic visual effects. Here, we also chose the perceptual metrics LPIPS and VGG distance. These representative methods and their quantitative comparisons are presented in Table \ref{tab3}. Note that all these results are from the respective papers.

\begin{figure*}[htbp]
	\centering
	\subfigure{
		\begin{minipage}[t]{0.125\linewidth}
			\centering
			\includegraphics[width=0.8in]{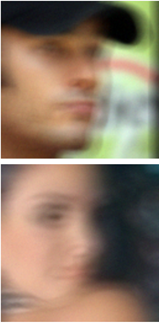}
			%\caption{fig1}
		\end{minipage}%
	}%
	\subfigure{
		\begin{minipage}[t]{0.125\linewidth}
			\centering
			\includegraphics[width=0.8in]{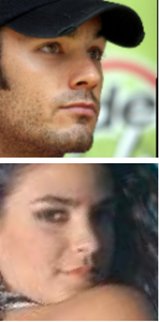}
			%\caption{fig2}
		\end{minipage}%
	}%
	\subfigure{
		\begin{minipage}[t]{0.125\linewidth}
			\centering
			\includegraphics[width=0.8in]{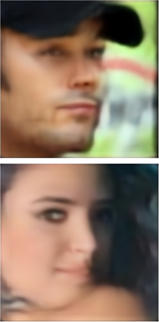}
			%\caption{fig2}
		\end{minipage}
	}%
	\subfigure{
		\begin{minipage}[t]{0.125\linewidth}
			\centering
			\includegraphics[width=0.8in]{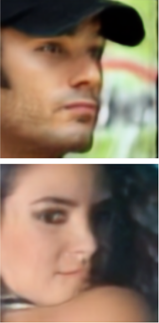}
			%\caption{fig2}
		\end{minipage}
	}%
	\subfigure{
		\begin{minipage}[t]{0.125\linewidth}
			\centering
			\includegraphics[width=0.8in]{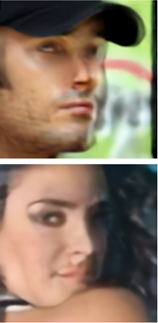}
			%\caption{fig2}
		\end{minipage}
	}%
	\subfigure{
		\begin{minipage}[t]{0.125\linewidth}
			\centering
			\includegraphics[width=0.8in]{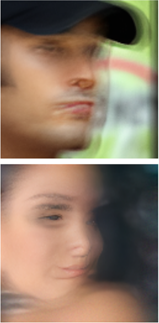}
			%\caption{fig2}
		\end{minipage}
	}%
	\subfigure{
		\begin{minipage}[t]{0.125\linewidth}
			\centering
			\includegraphics[width=0.8in]{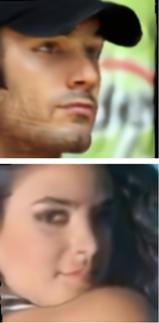}
			%\caption{fig2}
		\end{minipage}
	}%
	\centering
	 
	\caption{Qualitative comparisons of the representative deep learning methods on the Shen test dataset \cite{shenDeepSemanticFace2018}.   From left to right: blurry image, ground truth, Shen et al. \cite{shenDeepSemanticFace2018}, UMSN \cite{yasarla2020deblurring}, MSPL \cite{leeProgressiveSemanticFace2020}, Wang et al. \cite{wang2021towards}, and DFPG \cite{jungDeepFeaturePrior2022}.}
	\label{fig14}
\end{figure*}

Shen et al. \cite{shenExploitingSemanticsFace2020} did not provide their model implementation and their results on perceptual metrics in their paper, and, therefore, we do not show them here. For pixel-level evaluation metrics, such as PSNR and SSIM, the UMSN and DFPGnet models achieved the best and second-best results, respectively. This shows that, when used properly, both semantic segmentation maps and generative priors can help the model to achieve good performance. For the perceptual metrics such as $d_{VGG}$ and LPIPS, the DFPGnet and MSPL models achieved the best and second-best results, respectively. This shows that the network based on a generative prior can effectively learn the perceptual similarity between images. In addition, the unsupervised method proposed by Xia et al. \cite{xia2019training} achieved excellent performance next only to those of the above methods and outperformed most unsupervised methods. This model requires two images of the same scene with different degrees of blur as inputs to compensate for the lack of ground truth. This setting can help in the training process of the model and make it more robust. That is, adding auxiliary images to the unsupervised model is an idea that is worth trying. In contrast, unpaired learning methods \cite{luUnsupervisedDomainSpecificDeblurring2019} and multi-task blind face restoration methods \cite{wang2021towards} achieved poor results.

\begin{figure*}[htbp]
	\centering
	\subfigure{
		\begin{minipage}[t]{0.1\linewidth}
			\centering
			\includegraphics[width=0.7in]{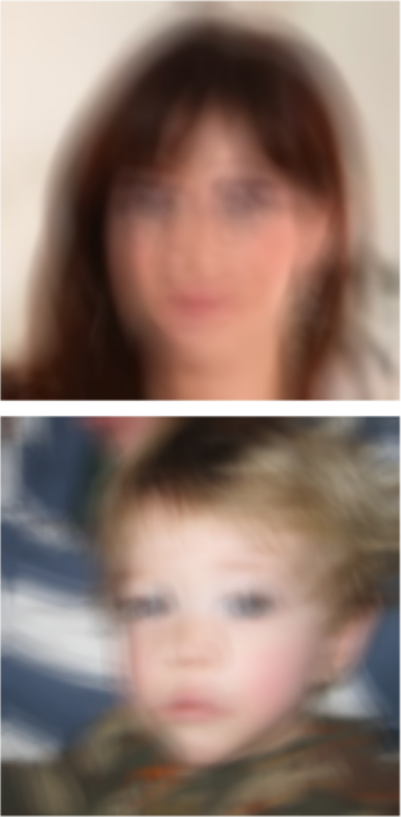}
		\end{minipage}%
	}%
	\subfigure{
		\begin{minipage}[t]{0.1\linewidth}
			\centering
			\includegraphics[width=0.7in]{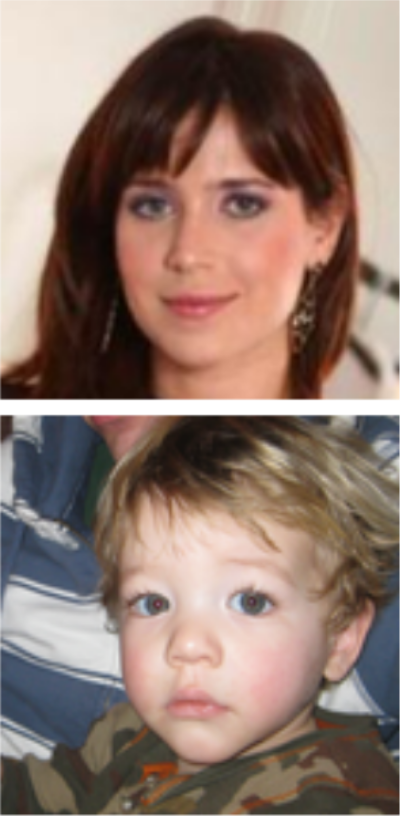}
		\end{minipage}%
	}%
	\subfigure{
		\begin{minipage}[t]{0.1\linewidth}
			\centering
			\includegraphics[width=0.7in]{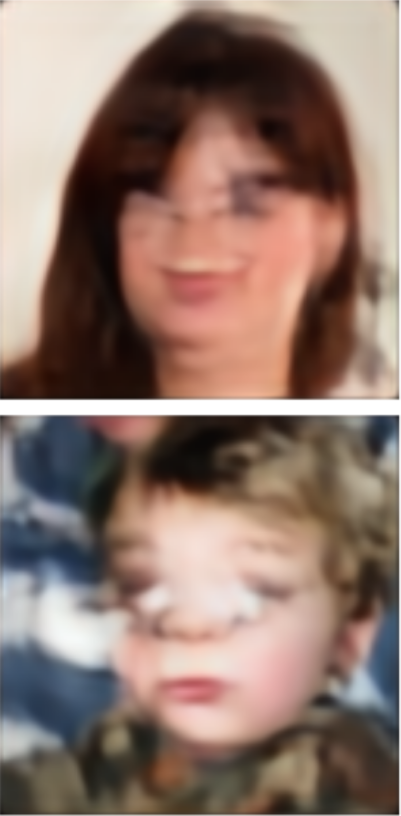}
		\end{minipage}
	}%
	\subfigure{
		\begin{minipage}[t]{0.1\linewidth}
			\centering
			\includegraphics[width=0.7in]{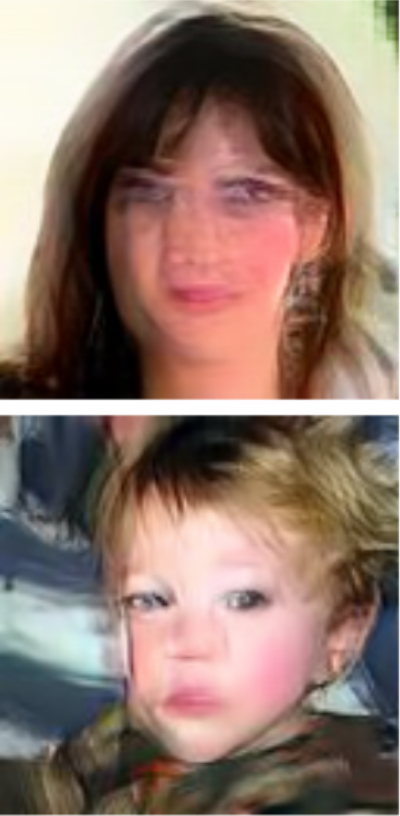}
		\end{minipage}
	}%
	\subfigure{
		\begin{minipage}[t]{0.1\linewidth}
			\centering
			\includegraphics[width=0.7in]{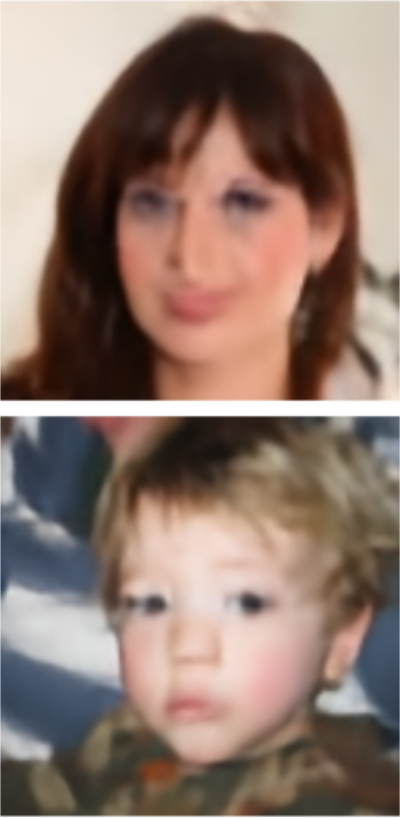}
		\end{minipage}
	}%
	\subfigure{
		\begin{minipage}[t]{0.1\linewidth}
			\centering
			\includegraphics[width=0.7in]{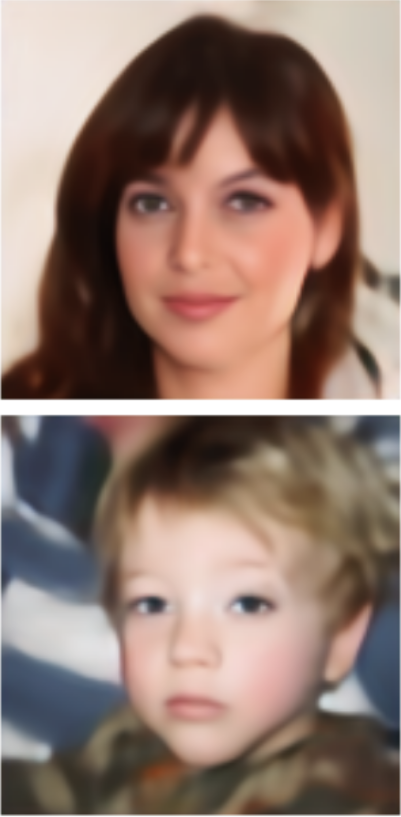}
		\end{minipage}
	}%
	\subfigure{
		\begin{minipage}[t]{0.1\linewidth}
			\centering
			\includegraphics[width=0.7in]{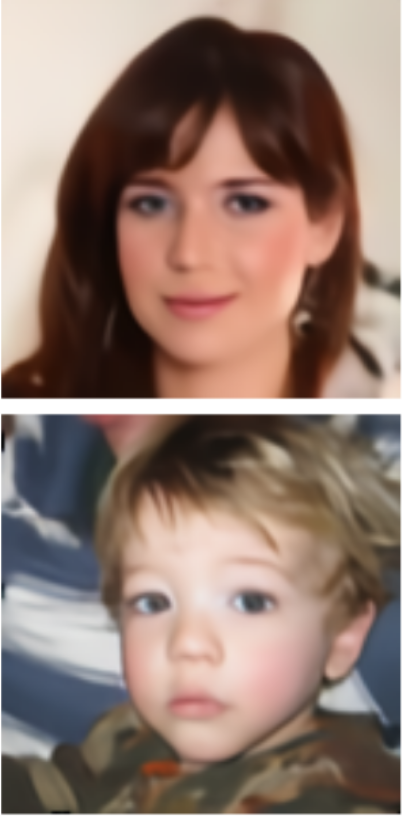}
		\end{minipage}
	}%
	\subfigure{
		\begin{minipage}[t]{0.1\linewidth}
			\centering
			\includegraphics[width=0.7in]{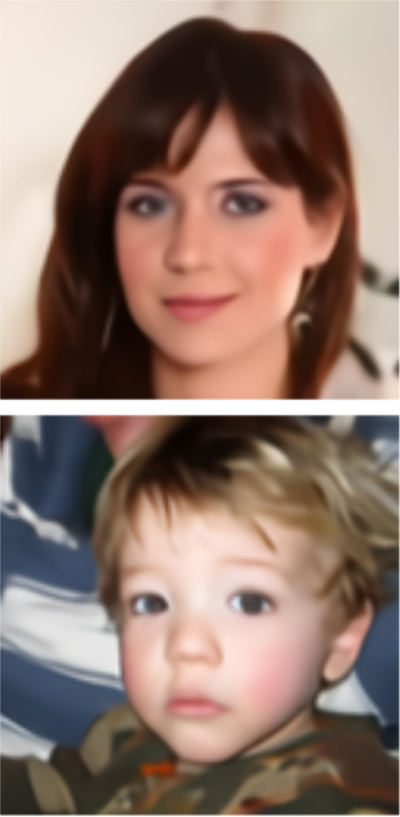}
		\end{minipage}
	}%
	\centering
	 
	\caption{Qualitative comparisons of the representative deep learning methods on the MSPL test dataset \cite{leeProgressiveSemanticFace2020}. From left to right: blurry image, ground truth, Shen et al. \cite{shenDeepSemanticFace2018}, Lu et al. \cite{luUnsupervisedDomainSpecificDeblurring2019}, Xia and Chakrabarti \cite{xia2019training}, UMSN \cite{yasarla2020deblurring}, MSPL \cite{leeProgressiveSemanticFace2020}, and DFPG \cite{jungDeepFeaturePrior2022}.}
	\label{fig15}
\end{figure*}

We also present the qualitative comparison between the Shen et al. dataset and MSPL dataset in Figs.~\ref{fig14} and ~\ref{fig15} which was given by \cite{jungDeepFeaturePrior2022}. Shen et al. \cite{shenExploitingSemanticsFace2020} introduced the semantic information of the face as a prior to the network for the first time and achieved a breakthrough deblurring result. However, it produces overly smooth results with unnatural restoration results in small areas such as the eyes and nose. Models such as UMSN \cite{yasarla2020deblurring} assign different weights to face components, providing clear results even for small face components. This shows that introducing a semantic segmentation map into the model and processing different semantic regions separately can improve the performance of the model. However, the model cannot adequately recover the blur of the image background and the generated faces are somewhat distorted in identity. The representative method DFPGnet, which takes the deep features of the face as a prior, produces deblurring results that are consistent with the identity of the original image. This shows that the deep features of the image simultaneously contain information such as identity, shape, and texture, which can guide the network to generate more realistic results. However, in the facial beard, eyelashes, hair, and other parts, the models still failed to restore sharp results. As shown in Fig.~\ref{fig14}, the blind face restoration method \cite{wang2021towards} produces severe artifacts in the test images. This is because the synthetic training datasets that they used only considered a simpler Gaussian blur. Therefore, their method can only handle simple and small blurs. The method exhibited poor results when the degradation of the image was severe. As shown in Fig.~\ref{fig15}, even the best unsupervised methods cannot produce results comparable to those of supervised methods. Restoring sharp images without an aligned ground truth remains a difficult problem.

\section{Discussion}
In this section, we will discuss the main differences between model-based and learning-based methods. 
	
\textbf{Flexibility of the model.} Model-based methods usually need to manually design specific priors,  which are very important for the final deblurring performance to limit the solution space of the problem. In other words, we often need to design a more refined prior for specific requirements, which is very inflexible. In contrast, deep neural networks can fit complex and varied blur processes owing to their powerful fitting capability. Therefore, most of the approaches focus on designing more powerful network structures. However, this often requires retraining the network when there are differences in the image distribution. 

\textbf{Deblurring performance.} Traditional methods usually introduce strong assumptions, such as the existence of a large amount of edge information in a scene. These assumptions result in poor generalization of the model and in ringing and artifacts on severely degraded real blurry images. Benefiting from large model capacity and large-scale datasets, deep learning methods can learn a wide range of blurry patterns and produce better results than traditional methods on test images.

\textbf{Inference speed.} Traditional methods are based on an iterative optimization process; therefore, it takes several minutes to tens of minutes to process one image. In contrast, deep learning methods require only a few milliseconds to process an image during the test time.

In summary, learning-based methods outperform traditional methods in terms of flexibility, performance, and inference speed. Therefore, the current mainstream methods are all based on deep learning and are committed to exploring better network architectures or better training tricks.

\section{Limitations and future research}
Deblurring of facial images is a challenging research topic. Here, we summarize the problems and limitations of existing methods and provide possible future research directions.

\subsection{Model structure design}
 There are four main limitations to the current methods. First, most of the methods can only process spatially uniform blur. In real-world settings, many scenes have non-uniform blur, such as blur caused by human motion while the background is clear and blurry background but clear foreground caused by using a large aperture of the camera. Most of the existing methods are unable to deal with such spatially non-uniform blur, and, therefore, identifying the specific blurring degrees of different parts of an image is a problem that needs to be addressed. Second, the current methods are effective in frontal facial images. However, for side faces, facial images with partial occlusion (such as glasses and hands covering the face), and monitoring equipment images with multiple faces in one image, the result will be severely distorted and anamorphic. Third, when the motion blur is severe, current methods often fail. For severely blurry facial images, these methods can lead to issues such as displacement of the facial components.  Fourth, the recognition accuracy of deblurred facial images is still inferior to that of ground-truth images. The ultimate purpose of deblurring is to improve the accuracy of high-level visual tasks. At present, the most advanced methods have achieved the same accuracy in face detection as that for clear images. However, in terms of face recognition accuracy (calculated using MobileNet trained with ArcFace loss), there is still a gap of four percentage points between SOTA deblurred images and clear images \cite{jungDeepFeaturePrior2022}.
 
For problems 1 and 2, we can aim to design finer and better network structures. Many deblurring methods developed for general scenarios have designed models specifically for spatially nonuniform blur \cite{zhang2018dynamic,gao2019dynamic,yuan2020efficient}, which can be used as a reference. To improve the performance of various facial poses (front or side), we need to explore how to integrate various facial priors, such as semantic segmentation maps, generative priors, and 3D priors, into the design of network structures to maximize their effects. For problems 3 and 4, we can try to increase the model capacity to improve the fitting ability. Currently, many transformer-based structures \cite{zamir2022restormer,chen2021pre} have been developed for image restoration tasks. With the powerful fitting ability of the transformer, the performance of severely blurry images can be improved.

\subsection{Construction of datasets}
Creating aligned blurry-sharp face image pairs is very difficult. According to this survey, most of the existing datasets are generated by convolving the sharp images with predefined blur kernels. There are three main problems here. First, synthetic datasets cannot represent the distribution of real images in the wild. Second, most of the synthesized blur kernels are spatially invariant, which cannot represent dynamic scenes with different blurs in different places. Third, some methods perform better than others only on specific datasets, but the performance does not hold when switching to another dataset. Therefore, it is necessary to develop benchmark datasets that cover different blur types.

\subsection{Unsupervised learning} Supervised deep learning methods require a large amount of paired training data to provide supervised information for the model, resulting in data dependence. However, unpaired learning can be trained on real blurry data without a ground truth, thereby improving the reconstruction ability of the model on real blurry facial images. Therefore, it is necessary to develop unsupervised methods for face deblurring. 

Existing unsupervised methods are mainly based on a GAN. Because there are no ground-truth images for the training, it is necessary to design proper loss functions to overcome this disadvantage. Moreover, these methods do not fully utilize the semantic or structural information of faces for guidance. When there is no corresponding ground-truth image, the extracted semantic information tends to be inaccurate. Therefore, extracting the correct semantic information from blurry images and incorporating it into the training of the network are a potential direction for unsupervised learning.

\subsection{Model generalizability} The performance improvement of deep learning methods mainly relies on large-capacity models and large datasets. Large-capacity models are extremely computationally expensive during training, and their training results are overly dependent on the training dataset. There are gaps in the distribution of data for people with different skin colors, ages, and ethnicities. Owing to this gap between domains, deep learning models exhibit poor generalization performance across different datasets and are prone to artifacts and other phenomena. Therefore, methods for domain adaptation should be developed and implemented to address this problem.

\subsection{Computational cost} When deeper networks are used to improve performance, model parameters and computational complexity are also improved. It is difficult to deploy the large-capacity models on mobile phones or embedded devices of monitoring systems. Therefore, the development of a lighter face deblurring model that combines and utilizes the characteristics of the device such as a dual camera \cite{lai2022face} is a potential future direction. 
\appendix
\subsection*{Declaration of competing interest}
The authors have no competing interests to declare that are relevant to the content of this article.\\

% for bibtex
\bibliographystyle{CVMbib}
\bibliography{facedeblur}

\end{document}